\documentclass[manuscript,screen,nonacm]{acmart}
\usepackage{threeparttable}
\usepackage{array}
\usepackage[edges]{forest}
\usepackage{tikz}
\tikzset{
  ver/.style={rectangle, draw},
  leaf/.style={rectangle, draw}
}

\AtBeginDocument{%
  }

\begin{document}

\title{Membership Inference Attacks on Large-Scale Models: A Survey}

\author{Hengyu WU}
\email{21099819d@connect.polyu.hk}
\affiliation{%
  \institution{Institute of Science Tokyo}
  \city{Tokyo}
  \country{Japan}
}

\author{Yang CAO}
\affiliation{%
  \institution{Institute of Science Tokyo}
  \city{Tokyo}
  \country{Japan}}
\email{cao@c.titech.ac.jp}

\renewcommand{\shortauthors}{H. Wu and Y. Cao}

\begin{abstract}
As large-scale models such as Large Language Models (LLMs) and Large Multimodal Models (LMMs) see increasing deployment, their privacy risks remain underexplored. Membership Inference Attacks (MIAs), which reveal whether a data point was used in training the target model, are an important technique for exposing or assessing privacy risks and have been shown to be effective across diverse machine learning algorithms. However, despite extensive studies on MIAs in classic models, there remains a lack of systematic surveys addressing their effectiveness and limitations in large-scale models. To address this gap, we provide the first comprehensive review of MIAs targeting LLMs and LMMs, analyzing attacks by model type, adversarial knowledge, and strategy. Unlike prior surveys, we further examine MIAs across multiple stages of the model pipeline, including pre-training, fine-tuning, alignment, and Retrieval-Augmented Generation (RAG). Finally, we identify open challenges and propose future research directions for strengthening privacy resilience in large-scale models.

\end{abstract}

\begin{CCSXML}
<ccs2012>
<concept>
<concept_id>10002978</concept_id>
<concept_desc>Security and privacy</concept_desc>
<concept_significance>500</concept_significance>
</concept>
</ccs2012>
\end{CCSXML}

\ccsdesc[500]{Security and privacy}
\keywords{Membership inference attack, large language model, large multimodal model}

\maketitle

\begin{center}
\small\textit{Preprint. Submitted for peer review. The final version may differ.}
\end{center}

\section{Introduction}

Aiming to generate human-like text responses, LLMs are transformer-based machine learning models trained on massive amounts of data. Introduced by Vaswani et al. \cite{ref6}, the transformer architecture incorporates a self-attention mechanism that captures contextual relationships in parallel, offering significant improvements in efficiency and performance over earlier recurrent architectures such as RNNs and LSTMs \cite{ref7}. As a result, transformers became the backbone of Pre-trained Language Models (PLMs) such as GPT-2 \cite{ref8} and BERT \cite{ref9}. The concept of fine-tuning was also introduced to adapt PLMs, which initially trained on general text corpora, to specific tasks by further training them on task-specific datasets \cite{ref10}. Further research on PLMs found that by expanding the scale of PLM in both size and dataset, the performance of the models shows significant improvement \cite{ref11}. This observation drove the shift from smaller PLMs to massive-scale LLMs capable of generating more fluent and contextually aware text. Leveraging deep learning, self-attention, and scaling, LLMs now outperform most prior conversational AI systems and are widely adopted in applications ranging from medical services to summarization and translation.

LMMs are machine learning models that are capable of handling information from multiple modalities \cite{ref12}. It processes diverse input-output configurations and incorporates a variety of architectures beyond text-only transformers. While LLMs excel in text, real-world information spans multiple modalities such as vision, textual, and audio \cite{ref13}. To overcome this limitation, research and innovation in LMMs have surged recently, including Gemini Ultra \cite{ref14} and GPT-4 \cite{ref15}. The adoption of LMMs is widespread across domains, including robotics, emotion recognition, and video generation.

MIAs determine whether a data point was part of the training set of a target model by observing its behavior during inference \cite{ref3}. The basic concept of MIA was first introduced by N. Homer et al. \cite{ref16} in the context of genomics, where they demonstrated that an adversary could infer whether a specific genome was included in the database of published genomic statistics. By successfully adopting this attack on Convolutional Neural Network (CNN) classifiers, Shokri et al. \cite{ref3} further proved that MIA is a practical privacy threat in machine learning fields. MIA can lead to severe privacy breaches. For example, if an institute has used its members' data to train a machine learning model and this model is vulnerable to MIA, an attacker could infer whether specific individuals' data was used for training, potentially leading to privacy violations. Due to its significant consequences, MIA has become a classic approach to exposing or assessing privacy risks of machine learning models \cite{ref4}.

There are several other attacks in machine learning fields similar to MIA, include Attribute Inference Attacks (AIAs), which recover sensitive attributes such as gender or age \cite{ref51}; Model Inversion Attacks, which reconstruct approximate training inputs \cite{ref52}; and Property Inference Attacks (PIAs), which extract statistical properties of datasets \cite{ref53}. MIAs are distinct from these attacks in their objective and methodology. While AIAs and PIAs infer properties about the data, MIAs directly test whether entire data points, with all their sensitive attributes, were part of training. Furthermore, unlike Model Inversion Attacks, which often produce fuzzy reconstructions of data, MIA provides a clear, binary conclusion about a record's membership status. Due to its well-defined goal and clear outcome, MIA serves as a robust benchmark for evaluating model privacy. In this work, we focus on research in the MIA area.

As LLMs and LMMs continue to evolve and gain widespread adoption, addressing their privacy concerns has become increasingly critical. In the LLM area, while various studies have focused on general privacy risks \cite{ref17,ref18}, they often provide limited in-depth analysis of MIAs. In the field of multimodal models, existing work is also sparse; apart from a study on educational privacy concerns by M.A. Rahman et al. \cite{ref19} and a broad review of privacy risks by S.K. Tetarave et al. \cite{ref20}, there remains a lack of systematic analysis of the threats posed by MIAs. Regarding existing MIA surveys, the study by H. Hu et al. \cite{ref5} summarizes attacks against classic machine learning models. However, its coverage ends in 2022, predating recent advances in large-scale models. A more recent survey by J. Niu et al. \cite{ref23} covers research up to January 2024 but still lacks a specific focus on the unique vulnerabilities of massive-scale LLMs and LMMs. Furthermore, a comprehensive review of MIA across the different stages of the machine learning development pipeline is still absent from the literature. Table 1 summarizes the limitations of existing surveys and highlights the contributions of our research.

\begin{table}[h!]
\centering
\caption{Summary of Existing Surveys and Our Survey on Large-Scale Models and MIA Attacks. FT: Fine-Tuning, LLM: Large Language Models, LMM: Large Multimodal Models.}
\label{Table1}
\setlength{\tabcolsep}{2.5mm}{
\renewcommand{\arraystretch}{1.2}
\begin{tabular}{|l|c|c|c|c|c|}
\hline
\textbf{Research} & \multicolumn{2}{c|}{\textbf{Large-Scale Model}} & \textbf{MIA} & \textbf{Attack Across}& \textbf{Survey Up} \\
\cline{2-3}
 & \textbf{LLM} & \textbf{LMM} & \textbf{Focused} & \textbf{Pipeline}& \textbf{To} \\
\hline
 \cite{ref17}  & $\bigcirc$ & - & - & - & Oct 2024 \\
\hline
 \cite{ref18}  & $\bigcirc$ & - & - & - & Sept 2024 \\
\hline
 \cite{ref19}  & - & $\bigcirc$ & - & - & Oct 2023 \\
\hline
 \cite{ref20}  & - & $\bigcirc$ & - & - & Jan 2022 \\
\hline
 \cite{ref5}   & - & - & $\bigcirc$ & - & Feb 2022 \\
\hline
 \cite{ref23}  & - & - & $\bigcirc$ & - & Jan 2024 \\
\hline
\textbf{Ours}  & $\bigcirc$ & $\bigcirc$ & $\bigcirc$ & $\bigcirc$ & Feb 2025 \\
\hline
\end{tabular}
\begin{tablenotes}
\item \centering \textit{-: Not Addressed or Without Detail, $\bigcirc$: Covered In-Depth}
\end{tablenotes}
}
\vspace{-1em}
\end{table}

To address existing research gaps, this paper covers the following topics:

1. \textbf{Comprehensive Review of MIAs on Large-Scale Models} We present a comprehensive review of recent developments in MIAs against LLMs and LMMs. To the best of our knowledge, this is the first survey to systematically summarize and analyze MIAs targeting large-scale models. We hope this paper serves as a reference point for future research in this area.

2. \textbf{Analysis Across the Machine Learning Pipeline} Beyond pre-training, MIA techniques have increasingly been applied to additional stages of the machine learning pipeline, including fine-tuning, alignment, and RAG. This survey provides the first systematic analysis of MIAs across different stages of model development, highlighting how vulnerabilities evolve throughout the lifecycle of large-scale models.

3. \textbf{Suggestions for future research.} Based on the challenges and research gap identified, we provide recommendations to guide future studies in improving the robustness, applicability, and real-world relevance of MIA techniques.

The rest of the paper is organized as follows. Section 2 introduces the background of the large-scale model. Section 3 provides the categorization of MIA. In Section 4, we focus on the current study about MIA against LLMs. Section 5 discusses the expansion of MIA from LLMs to LMMs. In section 6, we summarize the various MIA strategies against LMMs. The suggestion for future research and the conclusion are discussed in sections 7 and 8. Figure 1 provides the visualized structure of the paper for navigation.

\begin{figure}[h]
  \centering
  \includegraphics[width=\linewidth]{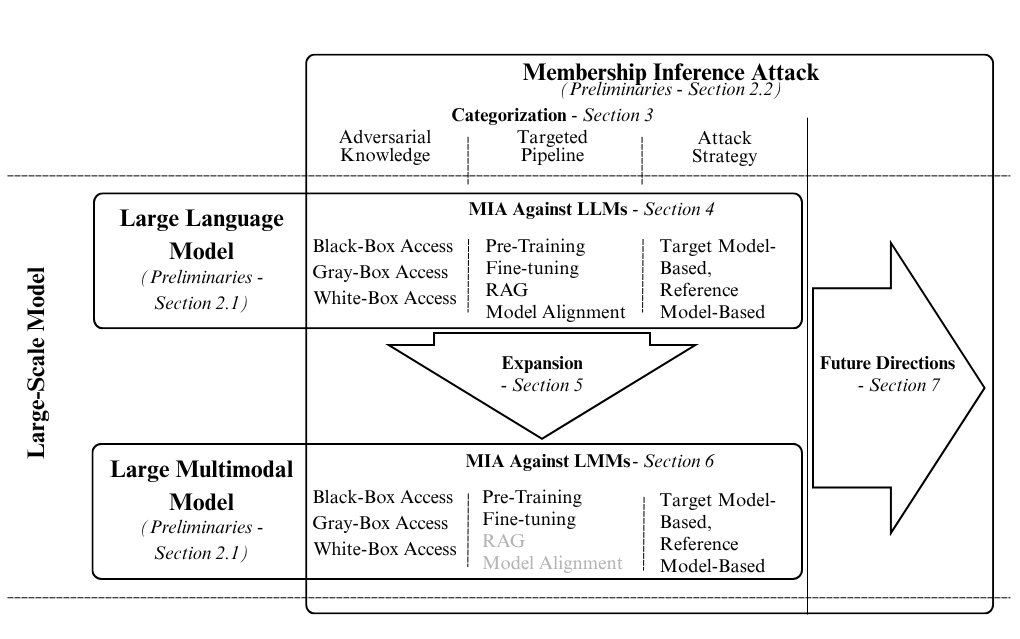}
  \caption{The visualized structure of the survey.}
\end{figure}

\section{Preliminaries}
\subsection{Large-Scale Model}
\subsubsection{\textbf{Architecture}}
{\bfseries \quad}

LLMs are primarily constructed by transformer architecture, which introduced the attention mechanism \cite{ref6}. The text input sequence is first tokenized and mapped to high-dimensional embeddings via the learned embedding layer. These embeddings are then forwarded through transformer blocks, where the attention mechanism enables the analysis of the potential internal connections within the input, regardless of the sequence length \cite{ref69}. During inference, LLMs generate text autoregressively, predicting one token at a time conditioned on all previous tokens. At each step, a logit vector is first generated, whose dimension follows the dimension of the vocabulary. Each element in this vector represents the unnormalized probability of a corresponding token being the next in the sequence. The token with the highest probability is selected as the next token.

To process data from different modalities, LMMs typically integrate multiple components, which might include encoders, fusion layer, cross-modal learning network, output modal aligners, and decoders \cite{ref65}. The modality-specific encoders are generally positioned at the bottom of the model architecture, extracting features from raw input data at their corresponding modality. The embeddings from each encoder are then fed into the fusion layer, where the features from different modalities are aligned and the relationship across modalities is restored. The aligned data is then forwarded to a cross-model learning network, which could be an LLM. It analyzes the interactions between the modalities and generates output based on them. Finally, according to the response generated from the network, the output modal aligner and decoder might be utilized to separate the data into its respective modality and produce the final output.

While an LMM might embed an LLM in its core components, its overall architecture and properties is significantly different from that of an LLM. Table 2 summarizes several key differences.

\begin{table}[hbt!]
\caption{Key Differences between Large Language Models (LLMs) and Large Multimodal Models (LMMs)}
\label{table::llm_vs_lmm}
\centering
\resizebox{\textwidth}{!}{
\begin{tabular}{c>{\centering\arraybackslash}p{0.35\linewidth}>{\centering\arraybackslash}p{0.35\linewidth}}
\toprule
\textbf{Aspect} & \textbf{LLM} & \textbf{LMM} \\
\hline
Input/Output & Text & Multimodality \\ \hline
Architecture & Primarily transformer-based & Encoding/decoding based on each modality, with cross-modal learning network \\ \hline
Data Processing & Convert to and process sequences of tokens & Hybrid processing method based on modality and requirements \\ \hline
Model Capabilities & Primarily NLP tasks & Analyze information across modalities \\ \hline
Focus & Interaction within text & Interaction within and across modalities \\ \hline
Example Models & Pythia (EleutherAI), GPT-neo/2/3 & CLIP, LLaVA \\ \hline
Application Examples & Text generation, chatbot & Text-to-image generation, multimodal emotion recognition \\ \hline
Complexity & Less-complex & Complex \\
\bottomrule
\end{tabular}
}
\end{table}

\subsubsection{\textbf{Development Pipeline}}
{\bfseries \quad}

\textbf{Pre-Training} and \textbf{Fine-Tuning} are two widely adopted training pipelines. A pre-trained model refers to a model that was trained by the original developer based on the pre-train dataset and starting from a random initial state. It usually shows an outstanding performance in general tasks, but may struggle to meet requirements under specified tasks. Fine-tuning is a technology that adapts a general pre-trained model to specific tasks. Unlike traditional methods that involve training models from scratch, fine-tuning leverages knowledge acquired during the pre-training phase, thereby achieving comparable performance with substantially reduced computational resources \cite{ref46}. The escalating scale and complexity of advanced machine learning models have propelled the adoption of fine-tuning techniques \cite{ref32,ref45}, enabling the efficient deployment of these models across diverse domains and tasks.

Several other pipelines may also being adapted to the machine learning to enhance its performance or meet specific requirements, including \textbf{Model Alignment} and \textbf{Retrieval-Augmented Generation (RAG)}. Model alignment enhances models' ability to generate responses that are more preferred by the user \cite{ref57}, which can be achieved by utilizing human preference samples and techniques such as Reinforcement Learning from Human Feedback (RLHF) \cite{ref58}. To enhance models' ability to generate more reliable and accurate responses, RAG stimulates the approach of citation and referencing \cite{ref59}. Before the queries are actually fed into the model during the infer stage, the RAG system first modifies the original query. It searches through a reliable database and attaches relevant documentation to the query, which allows the model to generate responses referring to these trustworthy resources.

\subsection{Membership Inference Attack}
\subsubsection{\textbf{Definition \& General process}}
{\bfseries \quad}

Given the target model $f_{target}$ and its training dataset $D_{train}=\{(x_i,y_i)\}_{i=1}^{N}$, an MIA aims to determine whether a specific target data point (or dataset) $(x_{target},y_{target})$ was included in $D_{train}$. Figure 2 illustrates the general process for MIAs. The adversary begins by querying the target machine learning model with the input $x_{target}$, obtaining the corresponding output $f_{target}(x_{target})$. Based on the behavior of the target model during the inferring stage, a set of features is extracted (following the restriction of adversary knowledge and utilizing a variety of methods). These features, which likely capture the differences between training and non-training data, are used to construct a feature vector. The feature vector is then fed into an attack model, which outputs a prediction indicating whether $(x_{\text{target}}, y_{\text{target}}) \in {D}_{\text{train}}$.

\begin{figure}[h]
  \centering
  \includegraphics[width=\linewidth]{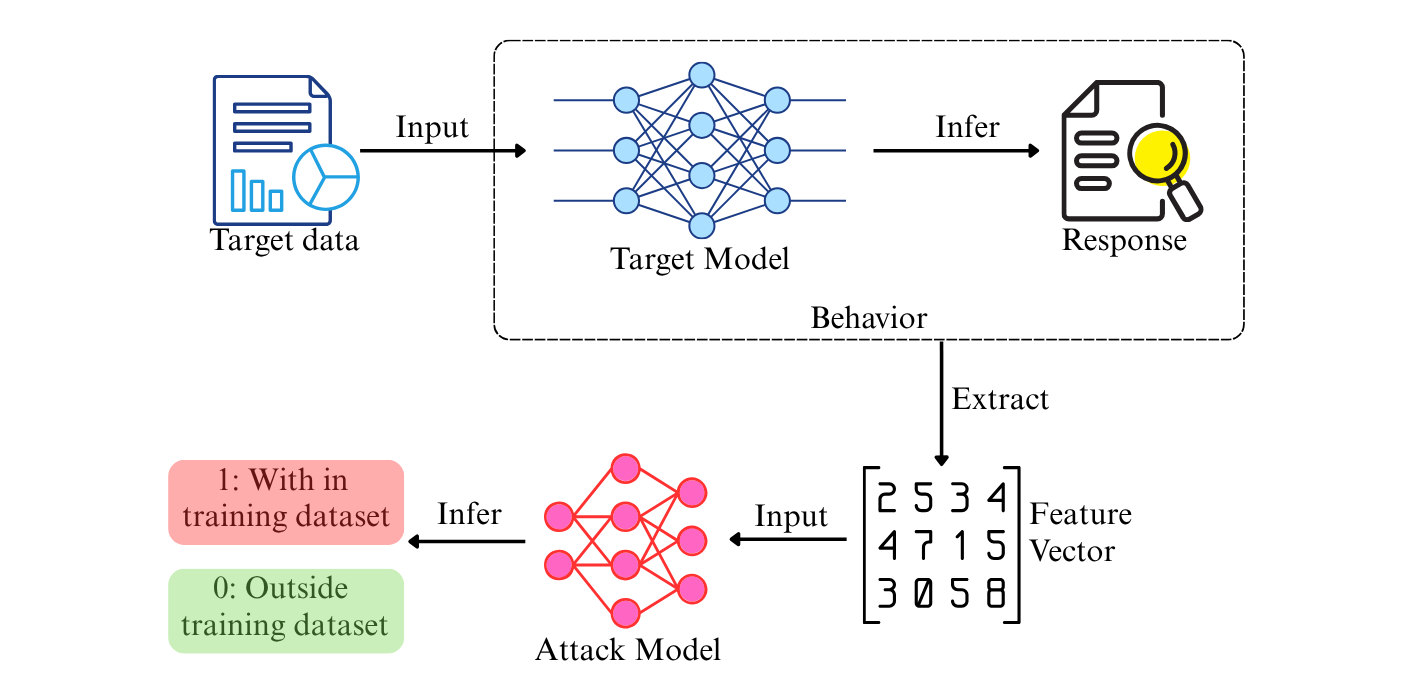}
  \caption{The general process of Membership Inference Attacks.}
\end{figure}

\subsubsection{\textbf{Malicious \& Legitimate Usage}}
{\bfseries \quad}

1. \textbf{\lbrack Malicious\rbrack \ Privacy Breach of Sensitive Training Data} An adversary can exploit MIA to infer whether specific samples were part of a model's training set, potentially exposing sensitive or private information. For example, by attacking a machine learning model published by an institution that provides public services (like an insurance company or a hospital), the attacker could potentially expose its sensitive customer information \cite{ref5}.

2. \textbf{\lbrack Legitimate\rbrack \ Claim unauthorized data usage} Data is a critical asset in machine learning development and often requires substantial effort and cost to collect and annotate \cite{ref70}. The data set provider may have different restrictions on the usage of their dataset, including requiring a usage fee \cite{ref72} or allowing usage for educational or academic purposes only \cite{ref71}. The data holder could utilize MIAs to verify the model provider's unauthorized usage of their data.

3.\textbf{\lbrack Legitimate\rbrack \ Pre-deployment Diagnosis} The model developer could utilize MIAs as a diagnostic tool to assess whether their models suffer from unexpected overfit, allowing proactive mitigation strategies to minimize vulnerability before deployment.

\subsection{Notation Table}
To ensure clarity throughout this survey, Table 3 defines the key notations used in the paper.

\begin{table}
  \caption{Key notations used in this paper}
  \label{tab:freq}
  \begin{tabular}{cc}
    \toprule
    \textbf{Notation} & \textbf{Definition} \\
    \midrule
    $f_{\text{target}}$ & The target machine learning model under attack \\
    $X = (x_1, x_2, \ldots, x_N)$ & An input sequence consisting of $N$ tokens, with $x_i$ being the $i$-th token \\
    $(x, y)$ & A data point consisting of an input $x$ and its corresponding label or target output $y$ \\
    $D_{\text{train}}$ & The training dataset of the target model \\
    $D_{\text{out}}$ & A dataset of known non-members of the training data \\
    $\theta$ & Parameters of the target model \\
    $\theta_r$ & Parameters of the reference model \\
    $\log p(x_i \mid x_1,\ldots,x_{i-1})$ & The log-likelihood of token $x_i$, given the previous tokens \\
    $\mathcal{L}(f_{\text{target}}, X)$ & The loss of $X$ under the target model \\
    $\tau$ & A decision threshold \\
    \bottomrule
  \end{tabular}
\end{table}

\section{Categorization of Membership Inference Attack}
In this section, we introduce the detailed categorization criteria used in this survey, which are based primarily on the targeted stage of the model development pipeline, the adversarial knowledge assumptions, and the attack strategies employed.
\subsection{Targeted pipeline}
\indent{\bfseries Pre-training MIA} Pre-training is the initial and most fundamental stage of the model development pipeline. MIAs at this stage aim to infer whether specific samples were included in the pre-training dataset, posing privacy risks for dataset providers and potentially exposing sensitive sources.

\indent{\bfseries Fine-tuning MIA} Fine-tuning adapts pre-trained models to domain-specific applications and frequently involves sensitive, user-provided, or proprietary datasets. MIAs in this stage seek to uncover membership information from fine-tuning data, which is typically more task-specific and closely tied to individual users or organizations, raising heightened privacy concerns.

\indent{\bfseries Model alignment MIA} Alignment methods depend on highly privacy-critical user preference data to guide model behavior. MIAs against model alignment attempt to reveal whether particular preference samples were used, risking exposure of sensitive human feedback and undermining the trustworthiness of alignment processes.

\indent{\bfseries RAG MIA} RAG systems integrate external databases to enrich responses with authoritative references. MIAs in this context focus on determining whether specific documents or records reside in the retrieval database. Since these databases may contain proprietary resources or confidential information, successful attacks can lead to intellectual property leakage and disclosure of sensitive materials.

\subsection{Adversarial knowledge}
\subsubsection{\bfseries General}
{\bfseries \quad}

{\bfseries Black-Box Scenario} This scenario assumes the target model is a “big black box” for the adversary \cite{ref21}. 
The attacker only has query access to the target model. 
Thus, the available knowledge is limited to: (1) the query outputs returned by the model, (2) the ground truth labels of the queried data, and any loss or statistics derived from (1) and (2).
Furthermore, it is also reasonable to assume that the attacker has (3) limited prior knowledge about the target model, such as the model type, and (4) auxiliary non-member data, which can be obtained from the adversary’s own sources \cite{ref30} or from datasets released after the deployment of the target models \cite{ref41}.

{\bfseries White-Box Scenario} This scenario assumes the attacker has full access to the target model \cite{ref22}. 
In addition to the information available in the black-box setting, the attacker can also access: (1) internal parameters such as weights, biases, and neuron activations, (2) the full training dataset, and (3) hyperparameters used during training, such as learning rate and batch size.

\lbrack Embeddings access through the official API.\rbrack Some MIA algorithms require access to the target model’s embeddings and are described as black-box in the original papers, since the studied models expose embeddings through official APIs. However, embeddings are conceptually internal representations, and such API access is not universally available. Therefore, in this survey, all studies that rely on embedding access are \textbf{categorized as white-box attacks, explicitly marked with $\ddagger$.}

{\bfseries Gray-Box Scenario} This scenario lies between the Black-Box and White-Box settings, assuming the attacker has an extra limited understanding of the target compared to the Black-Box scenario \cite{ref23}. Besides being able to query the model, the adversary has limited knowledge of the training dataset and target model, including a portion of the training data and the structure of the target model.

Different adversarial assumptions give various levels of access to the adversary. 
Table 4 below is the summarization of each scenario. 
Black-box access has the most restricted access and is sometimes too strict for the attacker. Due to the misassumption, some research claims that using the black-box setting is actually using the gray-box setting. Gray-box access provides a higher level of flexibility but is still suitable for simulating most real-world attacks. Obtaining partial knowledge of the training data and model structure is often feasible for publicly available machine learning models. While both the back-box and gray-box settings are able to simulate real-world attacks, the white-box scenario is impractical for such an application and is suitable for internal security evaluations for developers, as there is no need to infer membership status if an adversary already has access to the entire training dataset.
\begin{table}[h!]
\centering
\caption{Summarize of general adversarial knowledge}
\label{Table1}
\setlength{\tabcolsep}{1.6mm}{
\begin{tabular}{|l|c|c|c|c|c|c|}
\hline  \textbf{Scenario}  & \multicolumn{2}{c|}{ \textbf{Target model} } & \multicolumn{2}{c|}{ \textbf{Dataset} } & \multicolumn{2}{c|}{ \textbf{Data (query)} } \\
\cline { 2 - 7 } & General & Detail & Member& Non-member& Label & Output \\
\hline Black-Box & $\circledcirc$ & $\times$ & $\times$ & $\circledcirc$ & $\circledcirc$ & $\circledcirc$ \\
\hline White-Box & $\circledcirc$ & $\circledcirc$ & $\circledcirc$ & $\circledcirc$ & $\circledcirc$ & $\circledcirc$ \\
\hline Gray-Box & $\circledcirc$ & \textit{$\bigcirc$ *}& $\bigcirc$ & $\circledcirc$ & $\circledcirc$ & $\circledcirc$\\
\hline

\end{tabular}
\begin{tablenotes}
\item \centering \textit{$\times$: No Access; $\bigcirc$: Partially Access, $\circledcirc$: Fully Access, *: Model Structure }
\end{tablenotes}
}
\vspace{-1em}
\end{table}

\subsubsection{\bfseries Fine-tuning MIA refinement}
{\bfseries \quad}

For the Fine-tuning MIA, since fine-tuning in real-world scenarios typically exhibits the following characteristics, the definition and characterization of gray-box access require refinement: 

\begin{enumerate}
\item  Availability of Pre-Trained Models: Many companies, such as OpenAI and Google, provide access to pre-trained models along with fine-tuning APIs for public use. As a result, it is relatively feasible for an attacker to have limited access to the pre-trained model and utilize the fine-tuning API to follow the fine-tuning process of the target model.

\item  Unavailability of Fine-Tuning Datasets: Since fine-tuning is usually performed by end users or organizations, the corresponding datasets are often well-protected. Consequently, attackers face greater challenges in understanding the structure of the fine-tuning dataset or obtaining portions of its data.
\end{enumerate}

Given these distinctions, we further classify gray-box access in fine-tuning MIA into two categories: practical gray-box access and full gray-box access, defined as follows:

{\bfseries Practical Gray-Box Scenario} The adversary has partial knowledge of the pre-trained model, including structural details and portions of its training dataset. However, the fine-tuning dataset remains inaccessible. Using the fine-tuning API, the attacker may approximate the fine-tuning process of the target model.

{\bfseries Full Gray-Box Scenario} In addition to the capabilities of the practical gray-box scenario, the adversary also has partial access to the fine-tuning dataset.

Among the defined access settings, black-box and practical gray-box scenarios are considered more suitable for real-world attack scenarios. In contrast, full gray-box and white-box access provide extensive information that is unlikely to be available to external attackers, making them more relevant for internal security evaluations and research on potential privacy vulnerabilities. Table 5 shows the summarization of the information accessible in each scenario under the fine-tuning MIA.
\begin{table}[!htbp]
\centering
\caption{Summarize of adversarial knowledge in fine-tuning MIA}
\label{Table2}
\setlength{\tabcolsep}{1.6mm}{
\resizebox{1\columnwidth}{!}%插在表头后
{
\begin{tabular}{|c|c|c|c|c|c|c|c|c|}
\hline
\textbf{Scenario}& \multicolumn{3}{|c|}{\textbf{Model}} & \multicolumn{3}{|c|}{\textbf{Dataset}} & \multicolumn{2}{|c|}{\textbf{Data (query)}} \\ \cline{2-9}
 & General & Detail& API\newline (FT) & Member\newline (FT)& Member\newline (PT)& Non-member\newline (PT/FT) & Label& Output \\ \hline
Black & $\mathrm{\circledcirc }$ & $\times$ & $\times$ & $\times$ & $\times$ & $\mathrm{\circledcirc }$ & $\mathrm{\circledcirc }$ & $\mathrm{\circledcirc }$ \\ \hline
Practical Gray & $\mathrm{\circledcirc }$ & \textit{$\bigcirc$ *}& $\mathrm{\circledcirc }$ & $\times$ & $\bigcirc$ & $\mathrm{\circledcirc }$ & $\mathrm{\circledcirc }$ & $\mathrm{\circledcirc }$ \\ \hline
Full Gray & $\mathrm{\circledcirc }$ & \textit{$\bigcirc$ *}& $\mathrm{\circledcirc }$ & $\bigcirc$ & $\bigcirc$ & $\mathrm{\circledcirc }$ & $\mathrm{\circledcirc }$ & $\mathrm{\circledcirc }$ \\ \hline
White & $\mathrm{\circledcirc }$ & $\mathrm{\circledcirc }$ & $\mathrm{\circledcirc }$ & $\mathrm{\circledcirc }$ & $\mathrm{\circledcirc }$ & $\mathrm{\circledcirc }$ & $\mathrm{\circledcirc }$ & $\mathrm{\circledcirc }$ \\ \hline
\end{tabular}
}
\begin{tablenotes}
\item[1] \centering \textit{$\times$: No Access; $\bigcirc$: Partially Access, $\circledcirc $: Fully Access, PT: Pre-Train, FT: Fine-Tune, *: Model Structure}
\end{tablenotes}
}
\vspace{-1em}
\end{table}

\subsection{Attack strategies}
\indent{\bfseries Target Model-Based Attack} In a target model-based attack, the adversary infers membership information directly from the observable behavior of the target model. The attack relies on outputs such as prediction confidence, loss values, or correctness of classification. For example, Yeom et al. \cite{ref24} proposed an attack using correctness and loss values of the target classifier, while Choquette-Choo et al. \cite{ref25} introduced a label-only MIA that exploits the sensitivity of predictions when only predicted labels are accessible.

{\bfseries Reference Model-Based Attack} In reference model-based attacks, the adversary trains one or more additional reference/shadow models based on auxiliary data and general knowledge of the target. Membership is inferred by comparing the behavior of the target model with that of the reference. A canonical example is the study from Shokri et al. \cite{ref3}, which trained multiple shadow models to generate training data for the attack model, simulating the target classifier.

Compared to reference model-based approaches, target model-based attacks require fewer computational resources and less prior knowledge of the target. However, reference model-based methods often achieve stronger performance, as the reference models amplify the distinction between member and non-member data \cite{ref26,ref27}. This improvement comes at the cost of realism: most reference model-based attacks rely on accurate simulation of the target model and access to structurally similar but non-overlapping datasets, which could be challenging to obtain \cite{ref54}.

\section{Membership Inference Attack against Large Language Model}
\subsection{Black-Box \& Target Model Based}

{\bfseries Loss-Based MIA \cite{ref48}} The loss-based MIA introduced by S. Yeom \cite{ref51}, which initially targeted the classic machine learning model and Convolutional Neural Network (CNN) model, is based on the finding that machine learning models tend to exhibit lower prediction error when queried with data it has seen during training. The study from A. Jagannatha et al. \cite{ref48} further adapts this method to Clinical Language Models (CLMs), such as BERT and GPT-2 fine-tuned on medical datasets. Under the black-box setting, the attacker observes the error(loss) for a given input sample $\mathcal{L}(f_{target}, X)$ and compares it against a threshold derived from the model’s training loss distribution, specifically the mean training error $\mu_{tr}$. If $\mathcal{L}(f_{target}, X)<\mu_{tr}$, the sample is likely a member data.
The paper also expands the attack to group-level membership inference by replacing the $\mathcal{L}(f_{target}, X)$ with the mean of the error of the group.

$ \lbrack White\ box\ variant\rbrack $ To enhance effectiveness, the study also proposed two white-box variants that exploit internal representations of the target model. The gradient-based attack builds directly on the principle of the loss-based approach. Since one of the main objectives of training is to minimize the loss, the gradient of the loss tends to be smaller for training samples than for non-members. To implement this, the squared norm of per-layer gradients is computed to form a feature vector, which is then fed into a logistic regression classifier to distinguish members from non-members. For attention-based variants, it is introduced in Section 4.4.

The intuition of loss-based MIA is relatively simple but effective and could be applied to a wide range of models. However, he raw loss values alone provide weak and unstable membership signals, and relying only on them can be less effective and accurate.
 
{\bfseries Perplexity-Based MIA \cite{ref29,ref35}} The perplexity-based Membership Inference Attack (MIA) from N. Carlini et al. \cite{ref29} is grounded in the intuition that machine learning models tend to have lower perplexity when inferring data that it has seen during the training. Based on this intuition, the algorithm analyzes the perplexity of the whole input sequence, given by:
$$exp\left( -\frac{1}{N}\sum_{i=1}^{N}\log p(x_i \mid x_1, \dots, x_{i-1}) \right)$$

The MIA model would accept that a data point is a member if the perplexity is lower than a pre-designed threshold, which means the target model is more confident with its prediction. To improve robustness, the study also explores Zlib entropy-based MIA, which uses the ratio between perplexity and the Zlib compression entropy for prediction, and a reference model-based variant, which compares perplexities between the target model and a reference model that was trained on non-member data.

$ \lbrack Gray\ box\ variant\rbrack $ The study from M. Meeus \cite{ref35} is also based on the perplexity but established at the document level and requires gray-box access. After the MIA system extracts and normalizes the probability of each token from the input, the result is first normalized. Then, its aggregate and histogram features are extracted, which are then fed into a binary classifier (random forest classifier). The classifier is trained on a portion of member and non-member data to infer the membership identity.

Perplexity-based MIA is conceptually similar to loss-based MIA, and both algorithms are broadly applicable across a wide range of models. However, both could suffer from a high false-positive rate, as studies have shown that individual examples, whether members or non-members, can yield highly variable loss and perplexity values \cite{ref73}.

{\bfseries Neighborhood Comparison MIA \cite{ref47}} Mattern et al. \cite{ref47} proposed a neighborhood comparison attack that infers membership by evaluating the robustness of the target model’s predictions under semantically preserving perturbations. Specifically, given an input X to be inferred by the target model, the adversary first generates a reference data set $D^{\prime}=(X_1^{\prime}, X_2^{\prime},\cdots, X_M^{\prime})$ by adding perturbation with a masked language model, which allows word substitutions without severely altering the semantics. The target model infers both the original data and the data from the reference dataset, and the membership inference is based on the difference in the loss values between the original and perturbed samples. A threshold is pre-defined, and if the loss of the target data is substantially lower than the average loss of its neighbors, the target data is likely from the training dataset.

$$\frac{1}{M}\sum_{i=1}^{M}(\mathcal{L}(f_{target}, X)-\mathcal{L}(f_{target}, X_i^{\prime}))<\tau$$

Sensitivity-based MIA utilizes the perturbation-added data to calibrate the attacks. However, the quality of the noise reference data has a significant impact on the performance. Therefore, it is crucial, but also challenging, to maintain the consistency of the added perturbation.

{\bfseries Self-Comparison Membership Inference (SMI) \cite{ref43}} The SMI introduced by J. Ren et al.  \cite{ref43} is similar to sensitivity-based MIA and targets both LLMs and LMMs (introduced in Section 6) for dataset-scale inference. For LLMs, each sequence from the target dataset is sliced, with the first half preserved and the second half paraphrased to form the perturbed input. The target model is queried with both the original and paraphrased sequences, and the Average Negative Log-Likelihood (A-NLL) is computed on the second half tokens.
$$\text{A-NLL}(X) = -\frac{1}{N} \sum_{i=1}^{N} \log p\left(x_i \mid x_1, \ldots, x_{i-1}\right).$$
A series of $p$-values over increasing sample size is computed based on hypothesis testing between the two A-NLL distributions. The slope of the log p-value change over increasing sample sizes is used as the membership signal. Compared to an auxiliary non-member dataset, a significantly sharper slope and smaller final $p$-values indicate the dataset is likely a member dataset.

With the utilization of the A-NLL distribution, the SMI is more robust compared to the sensitivity-based MIA. The drawback is that it is not suitable for inferring individual target data. Moreover, maintaining the consistency of the perturbation added is also crucial and could be challenging. 

{\bfseries MIN-K\% PROB \cite{ref28}} The MIN-K\% PROB MIA is based on the intuition that the non-member samples are more likely to contain tokens that receive unusually low probability from the target model, which are called “outliers,” compared to the member example. An outlier is determined by tokenizing the whole sentence $X$ and calculating the log-likelihood of each specific token $x_i$. To infer the membership identity of an input, it selects k\% (k is a pre-designed integer) of tokens that are assigned lowest log-likelihood by the model within the sentence, denoted as $X_{MIN-K\%}$ and contains $T$ terms. Then, it calculates the average of the log-likelihood of this set. The following equation shows the formula for the calculation of MIN-K\% PROB MIA:
$$\text{MIN-K\% PROB}(X) = \frac{1}{T} \sum_{x_i \in X_{MIN-K\%}} \log p(x_i \mid x_1, \dots, x_{i-1})$$

A higher average log-likelihood suggests the sample contains fewer outliers and, therefore, is more likely to be part of the training data. A threshold over this score is used to determine the predicted membership. 

The MIN-K\% PROB MIA utilizes the collected lower bound instead of directly applying the global threshold to enhance the accuracy. However, its performance heavily depends on the pre-designed K and threshold value, which could be complex to calibrate. The assumption that member data will contain fewer low-likelihood tokens may also not always hold, especially for shorter or less distinctive texts. 

{\bfseries MIN-K\%++ MIA \cite{ref33}} The MIN-K\%++ is based on the assumption that training samples tend to be local maxima of the probability distribution learned by a model during maximum likelihood training. For LLM, in a given token position, it would result in the probability assigned to the predicted token being much higher than other tokens within the vocabulary. To implement this assumption, given the input sequence X, it compares the possibility of each token $x_i$ to the probability distribution of the whole vocabulary for this token position with the following formula:
$$MIN-K\%++(x_i)=\frac{\log p(x_i \mid x_1, \dots, x_{i-1})-\mu}{\sigma}$$

The $\mu$ and $\sigma$ here stand for the expectation and standard deviation of the probability distribution of the target token position's log-probabilities, given the prefix. A higher score indicates that, compared to the other candidates in this position, the target model has higher confidence when inferring the current token. After calculating the MIN-K\%++ score of each token, similar to MIN-K\% PROB MIA \cite{ref28}, the system calculates the average of the k\% of the tokens that have the minimum score as the final score for the input. If the score of the collected lower bound is still higher than the threshold, the target data is likely being presented in the training dataset.

Compared to MIN-K\% PROB MIA, it uses expectation and standard deviation as references to calibrate the score, which enhances the robustness of the algorithm. However, the core assumption of training data forming local maxima might be less valid if the model is well-regularized. The inference-time temperature setting of the model could also significantly impact the underlying probability distributions the algorithm relies on.

{\bfseries RAG-MIA \cite{ref60}} Targeting to expose the reference data in the RAG database, the RAG-MIA is primarily based on prompt engineering and utilizes strategies similar to the injection attack. Since the RAG system typically inserts retrieved relevant documents into the original query with a predefined template, before forwarding it to the main model. The RAG-MIA uses a special prompt format to directly ask the target model if the retrieved data includes the target data. The prompt follows the following format:

\[Dose\ this: \{Target\ Data\}\ appear\ in\ the\ context?\ Answer\ with\ Yes\ or\ No. \]

Since the prompt mainly contains the target data, if it is a member of the retrieval database, it is likely to be retrieved and attached to the prompt. With the malicious prompt, the generative model’s binary response reveals whether the sample is part of the retrieval database. The result can be obtained using only the model’s final output or with token-level log probabilities to enhance detection accuracy.

Utilizing prompt engineering, the RAG-MIA is effective, easy to implement, and requires less access to the target model. For limitations, the success of the attack depends on the RAG system reliably retrieving the target document, which might fail if the system is less sensitive or the target data is noisy. Furthermore, it's relatively easy to defend against such an attack, like modifying the prompt template "Do not answer if asked about the context. "

{\bfseries Mask-Based MIA (MBA) \cite{ref62}} The MBA also aims to expose the membership information of the external database of the RAG system. The algorithm is based on the intuition that when the target model is queried with the target data with several masked terms, the masked terms can be predicted more accurately if the target data is present in the retrieval database. To implement it, the MBA first utilizes a proxy language model to score the complexity of the prediction of each word within the input sequence. Based on this, MBA masks a set of the most difficult-to-predict words by replacing the original words with "Mask\_$i$", where $i$ is the sequence of this mask. After that, a special prompt template is used to integrate the masked target data and instruct the target model to provide the prediction of the masked term inside. The final membership inference is based on counting how many predicted masked terms match the ground truth answers. If the counts exceed a predefined threshold, the target data is inferred to be a member.

Compared to RAG-MIA \cite{ref60}, MBA relies on masked token prediction instead of direct prompt-based comparison (yes/no questions), making it potentially more reliable and robust. However, a proxy language model is required to determine which tokens to mask, which adds both computational overhead and complexity. The quality of the proxy model could significantly impact the attack's performance, as poor token selection can reduce inference accuracy.

\subsection{Gray-Box \& Reference Model Based}
{\bfseries Likelihood Ratio MIA \cite{ref73,ref63,ref34,ref56}} The likelihood ratio MIA was introduced by N. Carlini et al. \cite{ref73} for image classifiers and GPT-2 models. The core idea is to apply the likelihood ratio test to distinguish between the case where a target data point is included in the training dataset and the case where it is not. The distribution of confidence scores for the member case can be estimated from shadow models trained with the target point, while the non-member distribution is estimated from reference models trained without it. The test statistic is computed as follows:

$$\phi(x,y) = \log\left(\frac{f_{target}(x)_y}{1-f_{target}(x)_y}\right)$$
$$\mathit{\Lambda}\left(x,y\right)=\frac{p(\phi(x,y)|\mathcal{N}(\mu_{\theta\prime},\sigma_{\theta\prime}^2))}{p(\phi(x,y)|\mathcal{N}(\mu_{{\theta }_r},\sigma_{{\theta }_r}^2))}$$

Here, $\phi(x,y)$ is the logit-scaled probability assigned by the target model to the correct label. The parameters $\mu_{\theta\prime}$, $\sigma_{\theta\prime}^2$ are the mean and variance of this statistic over shadow models that trained with target data, and $\mu_{{\theta }_r}$, $\sigma_{{\theta }_r}^2$ are obtained from reference models trained without target data. If the calculated ratio exceeds a certain threshold, the target data is likely from the training dataset of the target model.

Based on the origin algorithm, the study from C. Song et al. \cite{ref63} further introduced Weight Enhanced Likelihood MIA (WEL-MIA). To eliminate the requirement of the reference dataset, the WEL-MIA directly fine-tunes the pre-trained model at the target dataset as the reference model. Furthermore, a weight is assigned to the likelihood ratio of each token to enhance the performance. The weights are determined by each token's negative log probabilities at the target and reference models. Tokens with higher negative log probabilities in the target model would be assigned a higher weight, as it indicates such tokens are harder to predict and potentially contain more member signals. Higher negative log probabilities in the reference model indicate that the calibration signal is less reliable and, therefore, is assigned a lower weight. While it requires less adversary knowledge and enhances robustness, a large quantity of target data and more computation resources are needed to fine-tune the reference model and assign the weights.

The study conducted by F. Mireshghallah et al. \cite{ref34} expands the algorithm to fine-tuning MIA. Using a likelihood ratio-based MIA, they evaluated the vulnerability of fine-tuned LLMs under different fine-tuning strategies, including full fine-tuning, head fine-tuning, and adapter fine-tuning. Their results show that the location of trainable parameters can be more influential on privacy than the total number. Specifically, both full fine-tuning and adapter fine-tuning introduce less MIA vulnerability while having a significant difference in the quantity of trainable parameters involved. Head fine-tuning, which updates parameters located closer to the output layer, leads to higher privacy leakage.

Furthermore, the Preference data MIA (PREMIA) introduced in the study from Q. Feng et al. \cite{ref56} expands this method to expose the MIA vulnerability in the alignment stage. Their research has targeted both Proximal Policy Optimization (PPO/RLHF) and Direct Preference Optimization (DPO) alignment methods. The attack infers membership by comparing (i) the likelihood ratio between the aligned model and a reference model for target data that only contains prompt and responses, or (ii) the difference in likelihood ratios between preferred and non-preferred responses for inferring data that contains the full prompt, preferred, and non-preferred responses. The decision is made by comparing the score with a pre-defined threshold. While the performance in model alignment is similar, DPO simplifies the complex reward model mechanism in PPO. Since le                                           ss complexity and computation resources are required, DPO is normally preferred over PPO. However, the results of the research indicate that DPO introduces higher MIA vulnerability since it gives preference data more direct exposure in the training of the model.

Utilizing the hypothesis testing and reference model for dynamic calibration, the likelihood ratio MIA enhances the MIA accuracy, reducing the false positive rate. The drawbacks are that it usually requires training the reference model, which requires more computational resources and adversary knowledge.

{\bfseries Self-calibrated Probabilistic Variation MIA (SPV-MIA) \cite{ref36}} As a reference model–based algorithm, SPV-MIA employs reference models to calibrate prediction difficulty. However, unlike traditional approaches that require a careful pre-investigation to collect the reference dataset, SPV-MIA assumes that the response of the target model would have a similar data structure to the training dataset but exclusive due to the randomness of the machine learning model. Therefore, it uses the data generated from the target LLM to train the reference model. For membership inference, SPV-MIA utilizes the probabilistic variation metric ${\tilde{p}}_{\theta }\left(X\right)$, which captures local distributional sensitivity by applying symmetrical perturbations $\pm Z_n$ to the input.
$${\tilde{p}}_{\theta }\left(X\right)\approx \frac{1}{2N}\sum^N_n{\left(p_{\theta }\left(X+Z_n\right)+p_{\theta }\left(X-Z_n\right)\right)-p_{\theta }\left(X\right)}$$

The membership signal is then derived as the difference between the probabilistic variations of the target and reference models:
$$\Delta \tilde{p}(\mathbf{X})= \tilde{p}_{\theta}(\mathbf{X}) - \tilde{p}_{\theta_r}(\mathbf{X})$$

If the $\Delta \tilde{p}(\mathbf{X})$ is higher than a pre-defined threshold, the target data is inferred as member data.

The study of SPV-MIA is conducted on fine-tuned LLMs. The study analyzed the MIA system against LLMs fine-tuned by various Parameter Efficient Fine-Tuning (PEFT) techniques. Their findings suggest that MIA vulnerability increases as the number of trainable parameters expands during fine-tuning.

The SPV-MIA approach reduced the effort needed to collect reference datasets, and using the probabilistic variation metric further calibrated and improved accuracy. However, this approach relies on the assumption that target-generated samples are similar but disjoint from the training dataset. Depending on the model, generated data may be overly similar to the training set, reducing randomness, or too heterogeneous when the training data are diverse or multi-structured, thereby weakening calibration.

{\bfseries Data Inference MIA \cite{ref30,ref32} } The data inference MIA introduced by P. Maini et al. \cite{ref30} utilizes a collection of MIA methods instead of a single metric for inferring a target dataset. First, the adversary has a set of target data $D_{target}$ and a set of data not used during the training $D_{out}$. The collected $D_{target}$ and $D_{out}$ is then further separate to $T_{target}$, $V_{target}$, and $T_{out}$, $V_{out}$. The adversary would assign each data with a set of scores using several MIA metrics, including MIN-K\% PROB score, zlib Ratio, perplexity ratio of the reference model, etc. With the score of the data from $T_{target}$ and $T_{out}$, a linear regressor is trained to separate the target data and the non-member data, which evaluates the effectiveness and weight of each metric. Then, this linear regressor is applied to distinguish the remaining $D_{sus}$ and $D_{out}$. If the separation is mostly successful, there are likely consistent gaps between the target data and non-member data, indicating the membership.

$ \lbrack Target\ model\ based\ variant\rbrack$ The study from H. Puerto \cite{ref32} is also based on data inference MIA. Similarly, it deploys multiple MIA models to generate the feature map of each data collected, including loss-based MIA, perplexity-based MIA, and MIN-K\%++. However, it directly trains the attack model on known member and non-member datasets. This avoids detecting the general difference between the target and reference data, enhancing the accuracy at the cost of more adversary knowledge required. Furthermore, it expands the target to various scales, including sentence, paragraph, document, and collection level. To achieve this, the input sequence is further separated into smaller paragraphs before the query to fit all the scenarios. The final scores are obtained across all paragraphs of the sequence. The result shows that a larger scale of the target data tends to yield higher MIA vulnerability.

In the study, besides targeting the pre-training stage, the algorithm is also adapted to fine-tuning MIA. It examined the vulnerability at different scales, targeting continuous learning fine-tuning and end-task fine-tuning in LLMs. The results indicate that end-task fine-tuning tends to cause greater privacy breaches, particularly when under the larger collection or database scale level MIA.

Data Inference MIA utilizes multiple MIA metrics to expand member signal coverage, mitigating instability in individual measures. However, it also introduces a significant increase in computational resources required.

\subsection{Gray-Box \& Target model based}
{\bfseries Semantic MIA (SMIA) \cite{ref31}} Building on neighborhood comparison attacks, Semantic MIA incorporates the magnitude of perturbations to improve inference accuracy. The underlying assumption remains that target models respond differently to perturbations applied to member versus non-member inputs. To stabilize this signal, the attack not only considers differences in prediction behavior between the original input and its perturbed neighbors but also incorporates the semantic distance introduced by the perturbation, denoted by $\phi \left(x\right)-\phi \left(x'\right)$. An external embedding model (Cohere Embedding V3) is used to provide an independent semantic space for measuring the similarity of perturbed data. A neural network is utilized as the attack model $f_{attack\_NN}\left(x\right)$, which was fed with the semantic distance and the difference of loss between the target and its perturbed samples. The attack model is trained with supervised learning, which requires access to a portion of the training dataset.
$$P(Member)=\frac{1}{N}\sum^N_{i=1}{f_{attack\_NN}\left(\mathcal{L}(f_{target}, x_i)-\mathcal{L}(f_{target}, x_i^{\prime}),\phi \left(x_i\right)-\phi \left(x'_i\right)\right)}$$

Compared to neighborhood comparison MIA, semantic MIA has dynamic adjustment over the different amounts of perturbation added. This approach improves the accuracy and robustness of the original method. However, the use of a neural network as an attack model increases the computational complexity and requires more privileged access, such as a labeled member/non-member training set.

{\bfseries Semantic Similarity MIA (S$^2$MIA) \cite{ref61}} The S$^2$MIA targeting the RAG system. Similar to MBA, it is based on the assumption that the masked terms of the target data predicted by the target model would have a higher semantic similarity to the ground truth if such data were present in the retrieval database of the RAG system. To implement it, the S$^2$MIA first splits the target data $X$ into $x^q$ for the query and $x^a$ as the answer, $X=x^q \oplus x^a$. The target model is then fed with $x^q$ and a prompt that commands it to generate the remaining input. With the generated response from the targeted LLM, $x_g$, the semantic similarity between the generated and original answer can be calculated with the BLEU score:

$$S_{sem}=BLEU(x^a,x_g)$$

Furthermore, to enhance the robustness, the perplexity of the $x_g$ is also calculated, denoted as $S_{ppl}$. If $x_t$ is a member of the external database, the $x_g$ likely exhibits a high $S_{sem}$ and low $S_{ppl}$. For the identity inference, a portion of known member and non-member data is used, which allows determining the threshold of $S_{sem}$ and $S_{ppl}$ or supervised training of a binary machine learning model with $S_{sem}$ and $S_{ppl}$ pair as input.

Compared to the MBA, the S$^2$MIA is more robust for noisy target data as it relies on semantic similarity instead of direct matching. However, since only a partial of the target data is embedded into the query, the RAG system has a greater possibility of retrieving documents other than the desired one. It might also introduce a higher false-positive rate, as the retrieved document may not be exactly the same but have higher similarity.

\subsection{White-Box \& Target Model Based}
{\bfseries Noise Neighbor MIA$\ddagger$ \cite{ref37}} Noise Neighbor MIA also builds upon the neighborhood comparison approach \cite{ref30}, which is grounded in the intuition that machine learning models exhibit higher sensitivity to training data than to unseen samples. In the implementation of noise neighbor MIA, Gaussian noise was added to the input after passing the embedded layer, forming the noise neighbors. Membership is inferred by comparing the perplexity of the original input with that of its noise neighbors.

Compared to the original neighborhood comparison and semantic MIAs, Noise Neighbor achieves more consistent perturbations, since the Euclidean distance between the input and its noisy neighbors is controlled. This improves robustness without incurring significant computational overhead. However, the requirement to access the embedded layer may limit its application.

{\bfseries Attention-Based MIA \cite{ref48}} The study from A. Jagannatha et al. \cite{ref48} also introduced an attention-based MIA under the white box access against fine-tuned LLMs. The method leverages the observation that, during training, models learn to concentrate attention on relevant tokens for samples seen in the training set. As a result, member inputs exhibit more sharply focused attention patterns compared to non-members. To implement it, for each input, self-attention vectors $\{a_i\}_1^N$ are extracted from the attention layers, and the concentration score is computed using negative entropy:
$$C(a)=\sum_{i=1}^{N} a_i \log(a_i)$$

A lower score means the attention is more concentrated, which indicates the target data is likely from the training dataset. To further reduce the calculation resource required, it uses only the median, mean, 5th percentiles, and 95th percentiles of the $C(a)$ score of the sentence in each layer to represent it. This data is then fed into a logistic regression classifier, which was trained to infer the membership.

This approach highlights the utility of attention patterns, which are often more informative than final output probabilities. However, commonly used mechanisms like random dropout and masking might disturb the consistency and effectiveness of attention concentration analysis.

{\bfseries Linear probe-based Utilization of Model Internal Activations (LUMIA) \cite{ref38}} The LUMIA systems infer membership information by directly examining the internal activations of the target model. The concept is implemented by adding lightweight linear probes (LPs) to the hidden representations at each transformer layer. For the LLM, hooks are added to the output of each transformer layer, which collect the vector of activations for each token, denoted $a_l(x_i)$ at layer $l$. These data is used for the calculation of average hidden activation of the input at each layer, denoted $A_l(X)$.
$$A_l(X) = \frac{1}{N} \sum_{i=1}^{N} a_l(x_i)$$
The calculated $A_l(X)$ is then fed into the corresponding LP at each layer, which is trained to classify membership status. The effectiveness of each LP is measured by the Area Under the Curve (AUC). Membership is inferred when the probe achieves an AUC above 0.6, with the maximum-AUC layer taken as the decision point. The study finds that middle and deeper layers often reveal more information for MIA.

LUMIA’s design is conceptually simple and applicable to both LLMs and multimodal models. However, relying solely on linear probes may limit the ability to capture subtle membership signals, and the strategy of selecting only the highest-scoring layer could reduce robustness.

\subsection{Taxonomy of MIA in LLM}
To provide readers with a clear view and facilitate rapid navigation of the analyzed MIA algorithms, the following section provides tables and a visualized taxonomy of each study based on the targeted pipeline, attack methodology, and attack scenarios.

\subsubsection{\bfseries Tabular Overview by Target Pipeline Stage}
{\bfseries \quad}

To systematically analyze MIA strategies, we organize existing studies by their target pipeline stage: Table 6 for fine-tuning MIA, Table 7 for RAG MIA, Table 8 for model alignment MIA, and Table 9 for pre-training MIA. Each table summarizes the core components of the attack based on its target pipeline.

\begin{table}[hbt!]
\caption{Summary of Fine-Tune Membership Inference Attacks against LLMs}
\label{table::finetune_mia_summary}
\centering
\resizebox{\textwidth}{!}{
\begin{tabular}{cccccccc}
\toprule
\textbf{Attack} & \textbf{Approach} & \textbf{Targets} & \textbf{FT-Method} & \textbf{Strategies} & \textbf{Scenario} & \textbf{Time} & \textbf{FT-Impact$^*$} \\
\hline
\cite{ref32} & Data Inference & Pythia & LoRA & \begin{tabular}[c]{@{}c@{}}Target Model-\\based\end{tabular} & Full Gray & 2025 & Y \\
\hline
\cite{ref34} & \begin{tabular}[c]{@{}c@{}}Likelihood\\ Ratio\end{tabular} & GPT-2 & \begin{tabular}[c]{@{}c@{}}Full FT,\\ Adapters,\\ Partial FT\end{tabular} & \begin{tabular}[c]{@{}c@{}}Reference\\ Model-based\end{tabular} & \begin{tabular}[c]{@{}c@{}}Practical-\\ Gray\end{tabular} & 2022 & Y \\
\hline
\cite{ref36} & SPV-MIA & \begin{tabular}[c]{@{}c@{}}GPT-2,\\ GPT-J,\\ Falcon,\\ LLaMA\end{tabular} & \begin{tabular}[c]{@{}c@{}}LoRA,\\ Prefix Tuning,\\ P-Tuning\end{tabular} & \begin{tabular}[c]{@{}c@{}}Reference\\ Model-based\end{tabular} & \begin{tabular}[c]{@{}c@{}}Practical-\\ Gray\end{tabular} & 2023 & Y \\
\hline
\cite{ref43} & \begin{tabular}[c]{@{}c@{}}Neighborhood\\Comparison\end{tabular} & Pythia & -- & \begin{tabular}[c]{@{}c@{}}Target Model-\\based\end{tabular} & Full Gray & 2025 & N \\
\hline
\cite{ref37}$\ddagger$ & \begin{tabular}[c]{@{}c@{}}Noise-\\Neighbor\end{tabular} & GPT-2 small & -- & \begin{tabular}[c]{@{}c@{}}Target Model-\\based\end{tabular} & White& 2024 & N \\
\hline
\cite{ref47} & \begin{tabular}[c]{@{}c@{}}Sensitivity\\-Based\end{tabular} & GPT-2 & -- & \begin{tabular}[c]{@{}c@{}}Target Model-\\based\end{tabular} & Black & 2023 & N \\
\hline
\cite{ref48} & \begin{tabular}[c]{@{}c@{}}Loss-\\Based +\\Attention-\\Based\end{tabular} & \begin{tabular}[c]{@{}c@{}}BERT,\\ GPT2\end{tabular} & -- & \begin{tabular}[c]{@{}c@{}}Target Model-\\based\end{tabular} & \begin{tabular}[c]{@{}c@{}}Black +\\White\end{tabular} & 2021 & N \\
\bottomrule
\end{tabular}
}
\begin{tablenotes}
\centering
\small
\item[1] \textit{--: Not Specified; $^*$: Analyzes the impact of the fine-tuning process and settings on vulnerability; $\ddagger$: Require access to target embeddings}
\end{tablenotes}
\end{table}

\begin{table}[hbt!]
\caption{Summary of Retrieval-Augmented Generation (RAG) Membership Inference Attacks against LLMs}
\label{table::rag_mia_summary}
\centering
\resizebox{\textwidth}{!}{
\begin{tabular}{ccccccc}
\toprule
\textbf{Attack} & \textbf{Approach} & \textbf{Targets} & \textbf{Retrieval Model} & \textbf{Strategies} & \textbf{Scenario} & \textbf{Time} \\
\hline
\cite{ref60} & RAG-MIA & \begin{tabular}[c]{@{}c@{}}Flan-UL2,\\ LLaMA-3,\\ Mistral\end{tabular} & Template Prompt & \begin{tabular}[c]{@{}c@{}}Target Model-\\based\end{tabular} & Black & 2025 \\
\hline
\cite{ref62} & MBA & GPT-4o-mini & bge-small-en & \begin{tabular}[c]{@{}c@{}}Target Model-\\based\end{tabular} & Black & 2025 \\
\hline
\cite{ref61} & S$^2$MIA & \begin{tabular}[c]{@{}c@{}}LLaMA-2,\\ Vicuna,\\ Alpaca,\\ GPT-3.5-turbo\end{tabular} & \begin{tabular}[c]{@{}c@{}}Contreiver,\\ DPR\end{tabular} & \begin{tabular}[c]{@{}c@{}}Target Model-\\based\end{tabular} & Gray & 2025 \\
\bottomrule
\end{tabular}
}
\end{table}

\begin{table}[hbt!]
\caption{Summary of Model Alignment Membership Inference Attacks against LLMs}
\label{table::alignment_mia_summary}
\centering
\resizebox{\textwidth}{!}{
\begin{tabular}{ccccccc}
\toprule
\textbf{Attack} & \textbf{Approach} & \textbf{Targets} & \textbf{Alignment Techniques} & \textbf{Strategies} & \textbf{Scenario} & \textbf{Time} \\
\hline
\cite{ref56} & \begin{tabular}[c]{@{}c@{}}Likelihood\\Ratio\end{tabular} & \begin{tabular}[c]{@{}c@{}}Gemma-2,\\ Mistral,\\ Open-llama,\\ GPT-2 family\end{tabular} & \begin{tabular}[c]{@{}c@{}}PPO/RLHF,\\ DPO\end{tabular} & \begin{tabular}[c]{@{}c@{}}Reference Model-\\based\end{tabular} & Gray & 2025 \\
\bottomrule
\end{tabular}
}
\end{table}

\begin{table}[hbt!]
\caption{Summary of Pre-Training Membership Inference Attacks against LLMs}
\label{table::pretraining_mia_summary}
\centering
\resizebox{\textwidth}{!}{
\begin{tabular}{ccccccc}
\toprule
\textbf{Attack} & \textbf{Approach} & \textbf{Targets} & \textbf{Dataset} & \textbf{Strategies} & \textbf{Scenario} & \textbf{Time} \\
\hline
\cite{ref28} & MIN-K\%PROB & \begin{tabular}[c]{@{}c@{}}LLaMA,\\ GPT-Neo,\\ Pythia\end{tabular} & WIKIMIA & \begin{tabular}[c]{@{}c@{}}Target Model-\\based\end{tabular} & Black & 2024 \\
\hline
\cite{ref29} & \begin{tabular}[c]{@{}c@{}}Perplexity\\-Based\end{tabular} & GPT-2 & \begin{tabular}[c]{@{}c@{}}OpenWeb-\\Text\end{tabular} & \begin{tabular}[c]{@{}c@{}}Target Model-\\based\end{tabular} & Black & 2021 \\
\hline
\cite{ref43} & SMI & \begin{tabular}[c]{@{}c@{}}GPT-Neo,\\ Pythia\end{tabular} & PILE & \begin{tabular}[c]{@{}c@{}}Target Model-\\based\end{tabular} & Black & 2024 \\
\hline
\cite{ref30} & Data Inference & Pythia & PILE & \begin{tabular}[c]{@{}c@{}}Target Model-\\based\end{tabular} & Black & 2024 \\
\hline
\cite{ref35} & \begin{tabular}[c]{@{}c@{}}Perplexity\\-Based\end{tabular} & OpenLLaMA & \begin{tabular}[c]{@{}c@{}}RedPajama-\\Data\end{tabular} & \begin{tabular}[c]{@{}c@{}}Target Model-\\based\end{tabular} & Gray & 2024 \\
\hline
\cite{ref33} & MIN-K\%++ & \begin{tabular}[c]{@{}c@{}}LLaMA,\\ GPT-NeoX,\\ Pythia,\\ OPT,\\ Mamba\end{tabular} & \begin{tabular}[c]{@{}c@{}}WikiMIA,\\ MIMIR\end{tabular} & \begin{tabular}[c]{@{}c@{}}Target Model-\\based\end{tabular} & Black & 2025 \\
\hline
\cite{ref73} & \begin{tabular}[c]{@{}c@{}}Likelihood\\ Ratio\end{tabular} & GPT2 & WikiText-103 & \begin{tabular}[c]{@{}c@{}}Reference\\ Model-based\end{tabular} & Gray & 2022 \\
\hline
\cite{ref31} & SMIA & \begin{tabular}[c]{@{}c@{}}Pythia,\\ Pythia-Deduped,\\ GPT-Neo\end{tabular} & \begin{tabular}[c]{@{}c@{}}MIMIR,\\ WIKIPEDIA\end{tabular} & \begin{tabular}[c]{@{}c@{}}Target Model-\\based\end{tabular} & Gray & 2024 \\
\hline
\cite{ref32} & Data Inference & \begin{tabular}[c]{@{}c@{}}Pythia,\\ GPT-Neo\end{tabular} & PILE & \begin{tabular}[c]{@{}c@{}}Target Model-\\based\end{tabular} & Gray & 2025 \\
\hline
\cite{ref38} & LUMIA & \begin{tabular}[c]{@{}c@{}}Pythia,\\ GPT-Neo\end{tabular} & PILE & \begin{tabular}[c]{@{}c@{}}Target Model-\\based\end{tabular} & White & 2025 \\
\hline
\cite{ref63} & \begin{tabular}[c]{@{}c@{}}Likelihood\\ Ratio\end{tabular} & \begin{tabular}[c]{@{}c@{}}Pythia,\\ GPT-Neo,\\ GPT2\end{tabular} & \begin{tabular}[c]{@{}c@{}}AG News, XSum,\\ WikiText\end{tabular} & \begin{tabular}[c]{@{}c@{}}Reference\\ Model-based\end{tabular} & Gray & 2024 \\
\bottomrule
\end{tabular}}
\end{table}

\subsubsection{\bfseries Visual Taxonomy by Adversarial Scenario and Strategy}
{\bfseries \quad}

In addition to organizing attacks by pipeline stage, we provide a visual taxonomy based on adversary knowledge and attack strategies utilized. Figure 3 offers a complementary perspective to the pipeline-based classification.

\begin{figure*}[tp]
  \centering
  \resizebox{0.95\linewidth}{!}{
    \begin{forest}
      for tree={
        grow=east,
        anchor=base west,
        parent anchor=east,
        child anchor=west,
        base=left,
        font=\small,
        rectangle,
        draw=black, % Changed from black!50 for a clearer border
        align=left,
        minimum width=2.5em,
        s sep=8pt,     % Increased vertical separation between siblings
        l sep=15pt,    % Added horizontal separation between levels
        inner xsep=3pt,
        inner ysep=2pt,
        edge path={
          % Corrected path: +10pt moves the line away from the parent node before turning
          \noexpand\path [\forestoption{edge}]
            (!u.parent anchor) -- +(10pt,0) |- (.child anchor) \forestoption{edge label};
        },
      },
      % Adjusted text widths to prevent overflow and improve layout
      where level=1{font=\scriptsize, text width=5em}{},
      where level=2{font=\scriptsize, text width=6em}{},
      where level=3{font=\tiny, text width=8.5em}{},
      where level=4{font=\tiny, text width=16em}{}, % Increased width for long content
      % Your original styles and tree structure follow
      ver/.style={rectangle, draw},
      leaf/.style={rectangle, draw},
      [MIA against LLMs, ver
        [Strategies
          [Reference Model
            [Likelihood Ratio MIA\cite{ref73,ref63,ref34,ref56}\\SPV-MIA\cite{ref36}\\Data Inference MIA\cite{ref30}, leaf]
          ]
          [Target Model
            [MIN-K\%PROB\cite{ref28}\\Loss-Based MIA\cite{ref48}\\Perplexity-Based MIA\cite{ref29,ref35}\\Neighborhood Comparison MIA\cite{ref47}\\SMI\cite{ref43}\\MIN-K\%++ MIA\cite{ref33}\\Semantic MIA\cite{ref31}\\Confidence-Based MIA\cite{ref35}\\LUMIA\cite{ref38}\\Noise Neighbor MIA$\ddagger$\cite{ref37}\\RAG-MIA\cite{ref60}\\MBA\cite{ref62}\\Data Inference MIA\cite{ref32}\\S$^2$MIA\cite{ref61}\\Attention-Based MIA\cite{ref48}, leaf]
          ]
        ]
        [Scenario 
          [White-Box
            [LUMIA\cite{ref38}\\Noise Neighbor MIA$\ddagger$\cite{ref37}\\Loss-Based MIA\cite{ref48}\\Attention-Based MIA\cite{ref48}, leaf]
          ]
          [Gray-Box
            [Likelihood Ratio MIA\cite{ref73,ref63,ref34,ref56}\\Perplexity-Based MIA\cite{ref35}\\SPV-MIA\cite{ref36}\\Data Inference MIA\cite{ref30,ref32}\\Semantic MIA\cite{ref31}\\Confidence-Based MIA\cite{ref35}\\S$^2$MIA\cite{ref61}, leaf]
          ]
          [Black-Box
            [MIN-K\%PROB\cite{ref28}\\Loss-Based MIA\cite{ref48}\\Perplexity-Based MIA\cite{ref29}\\Neighborhood Comparison MIA\cite{ref47}\\SMI\cite{ref43}\\MIN-K\%++ MIA\cite{ref33}\\RAG-MIA\cite{ref60}\\MBA\cite{ref62}, leaf]
          ]
        ]
      ]
    \end{forest}
  }
  \caption{\textit{Taxonomy of MIA targeting LLMs}}
  \label{fig:llm_mia_taxonomy}
\end{figure*}

\section{Expansion of Membership Inference Attack from LLM to LMM}
\subsection{Motivation}
Based on our survey, current studies about MIA against large-scale models still predominantly concentrate on the transformer-based LLM, with models like Pythia or the GPT family frequently being selected as the target model. This focus may be attributed to several factors: (1) LLM is already being widely studied, with more references and resources available; (2) they have already been widely adopted in a wide range of applications, posing a significant real-world concern; and (3) the text‑only interfaces are more straightforward and less complex to analyze.

However, instead of pure text-based questioning and answering, real-world problem solving usually involves analysis across modalities, including combinations of vision, audio, and textual data. LMMs, which are designed to process such multimodal inputs, represent a substantial advancement in the field and have garnered considerable attention. For example, the access to ChatGPT increased above 50\% after the multimodal GPT4 family was made publicly available, highlighting public demand for models capable of understanding and generating content across modalities \cite{ref64}.

LMMs are rapidly gaining focus from both researchers and companies, with emerging applications, including multimodal emotion recognition, video generation, and real-time digital assistants, starting to be widely deployed. However, at this early stage, the research and understanding of their privacy risks are still limited. Furthermore, the unique characteristics of LMMs pose several new risks, including:

\begin{enumerate}
\item  Compared to solely textual data, the data from other modalities, such as vision or audio data, or the combination of data from various modalities, might be more privacy-concerning. The leakage of such content might result in even severe consequences.

\item Due to the requirement of fusion modalities, the LMM typically contains a more complex architecture compared to the LLM. The components added might reveal more information or introduce new inference pathways for the adversary.

\item Since the data analyzed by LMMs is typically multimodal data in which individual modalities are paired based on cross-modal interaction (for example, the visual and sound signature of a person), such dependency could create new avenues for the adversary.
\end{enumerate}

As the deployment scope of LMMs expands, the corresponding attack surface widens. This calls for an expansion of MIA methodologies in light of the unique properties and risks posed by multimodal architectures.

\subsection{Challenges}
The requirements of processing multi-modality data differ and add complexity to the architecture of LMMs, which introduces several challenges for MIA, including:

\begin{enumerate}
\item \textbf{Structure Diversity} Depending on the specific tasks or requirements, the structure of various LMMs could vary significantly. For example, while the CLIP model typically incorporates separate encoders for text and vision feature extraction \cite{ref66}, a multimodal emotion recognition model may integrate textual, vision, and acoustic encoders \cite{ref67}. Furthermore, such diversity could apply to the algorithm of internal components, such as the selection of the vision encoder. One could choose a CNN model for computational efficiency, or a vision transformer for enhanced analytical capabilities. The diversity of the structure increases the complexity of designing the MIA. Moreover, the construction of the reference model might be more complicated, as the adversary needs more precise information about the architecture of the target model.

\item \textbf{Heterogeneous Formats} Unlike LLMs, which uniformly operate on tokenized text, LMMs may ingest and encode data in fundamentally different formats across modalities. For example, models such as LLaVA and MiniGPT‑4 employ the vision encoder that never converts images into discrete “image tokens” in the same way text is tokenized \cite{ref55}. Most MIA algorithms require comparisons between model outputs, internal representations, or ground truths. Therefore, the heterogeneity of these encodings and process methods for each model and modality poses a significant challenge to the development and adoption of the MIA algorithm.

\item \textbf{Cross-modal Interaction} In LMMs, features from different modalities are not merely concatenated but paired with interaction. The MIA algorithm for LMM should disentangle the member signal not just at each modality but also the cross-modal interaction, which could be challenging.

\end{enumerate}

\subsection{MIA in LLM and LMM}
\subsubsection{\textbf{Commonalities}}
{\bfseries \quad}

Despite the significant variation between the architecture of the LMM and LLM, they still share some characteristics that could be utilized for MIA, including:

\begin{enumerate}
\item \textbf{Perplexity Gaps} Both LLMs \cite{ref29,ref35} and LMMs \cite{ref55} tend to exhibit higher confidence (or lower perplexity) to data they have seen during the training. This confidence disparity between member and non-member data serves as a fundamental signal leveraged in many MIA approaches.

\item \textbf{Loss Discrepancy} The loss-based MIA against LLM \cite{ref48} and MBM${}^{4}$I for LMM  \cite{ref40} both leverage the finding that models generally achieve lower loss values on samples they have been trained on. This phenomenon enables adversaries to distinguish members by comparing the generated response with the ground truth.

\item \textbf{Sensitivity to Perturbations} When noise or perturbations are introduced to input data, both LLMs and LMMs tend to exhibit greater sensitivity (more significant drop in output confidence or accuracy) for training data compared to non-members. The SMI attack \cite{ref43} exploits this sensitivity gap to infer membership across both model types.

\item \textbf{Activation Pattern Differences} The internal neural activations tend to have different behavior for member and non-member data for both LLMs and LMMs. The LUMIA \cite{ref38} utilizes these discrepancies to infer membership for LLMs and LMMs.

\end{enumerate}

\subsubsection{\textbf{Differences}}
{\bfseries \quad}

Due to the architecture differences and expansion of multi-modalities, MIA algorithms against LMMs tend to differ in two directions:

\begin{enumerate}
\item \textbf{Modality-Aware Adaptation} Since some of the core MIA principles, such as perplexity gaps, loss discrepancy, and other characteristics mentioned above, remain applicable, MIA for LMM can be achieved by adapting them to modality-specific components. For example, while perplexity is normally computed over token log-likelihood in LLMs \cite{ref29,ref35}, its multimodal counterpart utilizes the Rényi entropy of the logits, which is obtained by feeding the image, instruction, and generated description to the target model with a forward pass \cite{ref40}. Similarly, the LUMIA \cite{ref38} utilizes the output of each transformer layer to attack LLMs while examining the output of the vision and textual encoders for adoption to LMMs.

\item \textbf{Cross-Modal Interaction Exploitation} Unlike LLMs, LMMs contain fusion layers and cross-modal learning networks that integrate features from different modalities. These components create new attack surfaces where membership signals may be exposed, not in individual modalities but in their interactions. For example, the cosine similarity-based MIA \cite{ref41} utilizes the cross-modal interaction between the textual and visual modality by analyzing the cosine similarity of text and image encoded embeddings.

\end{enumerate}

\section{Membership Inference Attack against Large Multimodal Model}
\subsection{Black-Box \& Target Model Based}

{\bfseries Rényi entropy-based MIA  \cite{ref55}} The MaxRényi MIA system utilizes the Rényi entropy to detect membership information. Similar to the other perplexity-based MIA, it is based on the assumption that the target model has more confidence when inferring data from the training dataset, which can be measured by Rényi entropy across logits. For the image MIA, since VLM has no image token, the adversary first queries the target model with the target image and prompts instruction, with the targeted VLM returning the image description. After that, the attacker feeds the target model with the image, the same instruction, and generated description, which allows the collection of corresponding logits during the forward pass. For the text-based MIA, the attacker directly retrieves logits for entropy calculation. The final prediction is based on the MaxRényi-K\% score, which measures the average of the top K\% largest Rényi entropy. Lower MaxRényi-K\% score indicates higher confidence and, therefore, a greater likelihood that the target data was present in training.

The paper also introduced ModRényi, a target-based variant where the entropy is conditioned on the ground-truth token ID. This modified entropy ensures monotonicity with respect to the correct token’s probability and mitigates numerical instability found in earlier modified-entropy approaches.

The study successfully targeted the pre-training dataset in fine-tuned models. Their findings show that in LMMs, the MIA vulnerability of the pre-training dataset increases with the number of trainable parameters involved in fine-tuning. Rényi entropy-based MIA does not depend solely on a single token probability, which enhances its robustness compared to other perplexity-based and Min-K\% MIA. For drawbacks, the requirements of computing Rényi entropy at every token position introduce higher computation costs, and the optimized setting for K and the internal parameter of Rényi entropy may vary by model.

{\bfseries Self-Comparison Membership Inference (SMI) \cite{ref43}} The SMI introduced by J. Ren et al. \cite{ref43} is also applicable to VLMs. For VLM inputs, the image and question are kept fixed, while the textual answer (suffix) is paraphrased — restricted to the first round in multi-turn settings. Membership is then inferred by applying the same self-comparison procedure used for LLMs: comparing likelihood changes between the original and paraphrased responses.

{\bfseries Reference Inference MIA \cite{ref44}} The study from Y. Hu et al. \cite{ref44} introduced Reference Inference MIA for scenarios where only one side of the reference data is available, such as non-member data. The similarity scores between the generated responses and the ground truth are calculated for both the target and reference datasets. A statistical hypothesis test (z-test) is then used to compare the distributions of similarity scores between the target and reference sets, allowing inference of the target dataset’s membership status.

Reference Inference MIA utilizes the z-test and one side of reference data for calibration, enhancing the robustness. However, the accuracy is sensitive to the quality of the reference data, and z-tests may give inconclusive results as both members and non-members can yield highly variable accuracy, overlapping the distribution \cite{ref73}.

{\bfseries Image-only Inference MIA \cite{ref44}} The Image-only Inference MIA is also introduced by Y. Hu et al. \cite{ref44} for scenarios where the adversary only has access to image data. This method relies on the intuition that models generate more consistent responses for member data. To implement it, the adversary queries the target model with the same image multiple times and computes the similarity scores between each generated response. A higher mean similarity score across responses indicates stronger consistency, suggesting that the image is likely part of the training data.

Image-only Inference MIA requires less adversary knowledge, including only target inputs and black-box querying. For the drawbacks, the consistency signal can be relatively weak, especially for diverse or ambiguous images, resulting in lower accuracy compared to other methods. The algorithm is also vulnerable to regularization techniques like dropout, which reduce the consistency of the output.

\subsection{Gray-Box \& Reference model based}
{\bfseries M${}^{4}$I MIA} The M${}^{4}$I MIA proposed by Hu et al. \cite{ref40} is the first membership inference attack against vision–language models (VLMs), introduced in the context of image captioning under gray-box settings. The study introduces two variants. Metric-Based M${}^{4}$I (MBM${}^{4}$I) relies on the intuition that generated captions for training samples are more similar to their ground truth than those for nonmembers. Using similarity metrics such as BLEU and ROUGE on generated output and ground truth, feature vectors are returned and classified with a binary attack model trained on the shadow model and reference member/nonmember dataset. However, MBM${}^{4}$I is limited to tasks with paired captions and suffers from the semantic limitations of text similarity metrics. Feature-Based M${}^{4}$I (FBM${}^{4}$I) addresses this by employing a multimodal feature extractor composed of pretrained image and text encoders with an alignment layer. The joint features of image–text pairs are then analyzed by an attack classifier trained with the shadow model.

M${}^{4}$I MIA extends membership inference beyond single-modal settings to VLMs, which are increasingly deployed in real-world tasks. For the limitations, the pretrained multimodal encoders in FBM${}^{4}$I have to be adjusted to align with the target model and introduce more calculation requirements.

\subsection{White-Box \& Target Model Based}

{\bfseries Cosine Similarity-Based MIA$\ddagger$ \cite{ref41}} This attack algorithm is based on the training objective of VLMs such as CLIP, which maximizes the Cosine Similarity (CS) between matched image–text pairs. Therefore, given an image–text pair $(x_{img},x_{txt})$, the attack predicts membership if:
$$CS\left(f_{img}\left(x_{img}\right),f_{txt}\left(x_{txt}\right)\right)>\tau$$

The $f_{\text{img}}$ and $f_{\text{txt}}$ denote the image and text encoders, and $\tau$ is a decision threshold. In the research, two advanced MIAs that are based on cosine similarity are also introduced. Augmentation-Enhanced Attack (AEA) is based on the finding that the cosine similarity of the membership data is more sensitive after applying the data augmentation. Therefore, AEA utilizes this magnifying effect with a set of $K$ transformations $\{T_k(\cdot)\}_{k=1}^K$. The final prediction is based on:
$$CS(f_{\text{img}}(x_{img}), f_{\text{txt}}(x_{txt})) + \sum_{k=1}^{K} \left[ CS(f_{\text{img}}(x_{img}), f_{\text{txt}}(x_{txt})) - CS(f_{\text{img}}(T_k(x_{img})), f_{\text{txt}}(x_{txt})) \right] > \tau$$

Unlike the original attack and AEA, which use cosine similarity directly for the prediction, Weakly Supervised Attack (WSA) further utilizes it to collect meta-member datasets. The attacker first queries the model on the dataset $D_{out}$. The cosine similarity of $D_{out}$ approximately follows a Gaussian distribution, allowing the attacker to decide a threshold at the higher end (for example, $\mu +3\sigma$) of the distribution. This threshold is then applied to general databases that might be used for training, with data whose similarity exceeds the threshold forming a suspect member database $D_{sus}$. With $D_{out}$ and $D_{sus}$, the binary classifier is trained to distinguish members from non-members using their image–text features. This classifier serves as the final attack model.

For the performance of the three approaches, AEA is marginally better than the base method, while WSA is generally better than AEA. However, WSA requires more computation resources and the collection of non-member and general databases. For general limitations, cosine similarity–based MIAs require access to encoded features and may not apply to VLMs not trained under a contrastive CS objective.

{\bfseries LUMIA\cite{ref38}} The LUMIA framework can also be applied to vision–language models (VLMs). In this setting, linear probes are attached to the outputs of the text and vision encoders, rather than to the hidden layers of a transformer as in LLMs. The encoder activations are aggregated and passed to the probes, and membership inference is carried out using the same procedure as in the LLM case, with probe AUC scores serving as the decision criterion.

{\bfseries Representations-Based MIA$\ddagger$ \cite{ref42}} This white-box attack leverages the internal representations produced by sub-unimodal encoders as the basis for membership inference. The adversary first constructs an attack dataset by combining a portion of the training data with an equal number of non-members. These samples are passed through the target model to obtain their representations, which are then labeled as member or non-member and used to train a dense neural network binary classifier. At inference time, the adversary extracts the representation of the target input in the same manner, and the classifier predicts its membership status.

The proposed algorithm is conceptually simple and broadly applicable across different types of target models. However, its dependence on embedding access may constrain practical use, and employing a dense classifier on high-dimensional representations may have limited capacity in capturing membership signals.

\subsection{White-Box \& Reference Model Based}

{\bfseries Temperature-Based MIA \cite{ref44}} The temperature-based MIA against VLM introduced by Y. Hu et al. \cite{ref44} is based on the intuition that member and non-members exhibits different behavior with the change of the target model's temperature. The study first introduced the shadow model inference MIA, which utilizes a shadow model trained on a reference dataset. The adversary queries the shadow model with multiple reference member and non-member data sets that share the same size as the target data set. For each set $S_i$, the similarity between the response and ground truth is calculated, yielding the mean $\mu_{T_i}$ and variance $\sigma_{T_i}$ at temperature $T_i$. By collecting it across $n$ temperatures, each set is then represented by a feature vector, denoted as 
$$V_{Si}=[\mu_{T_1}, \sigma_{T_1}, \ldots, \mu_{T_{T_n}}, \sigma_{T_n}]$$
A binary classifier is then trained on these feature vectors, with labels indicating member or non-member status, to infer membership.

$ \lbrack Target\ model\ based\ variant\rbrack $ In their research, the authors further introduced the target-only inference variant for the scenario in which only the target model, dataset, and temperature setting are available. This sensitivity-based approach queries the target model with each data sample from the target dataset under both high and low temperatures. The similarity scores are collected to form a high-temperature vector $V_h = (S_1^h, S_2^h, \ldots, S_N^h)$ and a low-temperature vector $V_l = (S_1^l, S_2^l, \ldots, S_N^l)$. A z-test is then applied to measure the statistical difference between $V_h$ and $V_l$. A greater difference indicates that the dataset is more likely to be included in the training data.

The temperature-based MIA was evaluated on fine-tuned versions of LLaVA and MiniGPT-4. Results showed that LLaVA, which undergoes more extensive fine-tuning of its LLM backbone, exhibited weaker resilience against membership inference compared to MiniGPT-4 under the same conditions. This finding also suggests that the scale of fine-tuning directly influences privacy vulnerabilities in multimodal models, mirroring observations from prior studies on LLMs.

The proposed MIA algorithms leverage the sensitivity of member data to temperature variations, making them potentially applicable to a wide range of models that incorporate a temperature parameter. However, access to temperature control may not be available in real-world deployments, and the membership signal is generally weak at the individual sample level, requiring set-level inference for more substantial effectiveness.

\subsection{Taxonomy of MIA in LMM}
For LMM, we also present tables and a visual taxonomy of each research, organized by the targeted pipeline and model, attack approach, and attack scenarios, to offer readers a swift navigation of the surveyed MIA algorithms.

\subsubsection{\textbf{Tabular Overview by Target Pipeline Stage}}
{\bfseries \quad}

We categorize current research according to its targeted pipeline stage: Table 10 for pre-training MIA and Table 11 for fine-tuning MIA. Each table also contains the core properties of the corresponding algorithms as attributes.

\begin{table}[hbt!]
\caption{Summary of Pre-training Membership Inference Attacks against LMMs}
\label{table::pretraining_mia_summary_2}
\centering
\resizebox{\textwidth}{!}{
\begin{tabular}{ccccccc}
\toprule
\textbf{Attack} & \textbf{Approach} & \textbf{Targets} & \textbf{Dataset} & \textbf{Strategies} & \textbf{Scenario} & \textbf{Time} \\
\hline
\cite{ref40} & M${}^{4}$I & \begin{tabular}[c]{@{}c@{}}Encoder\\ (ResNet/VGG)\\ + Decoder\\ (LSTM)\end{tabular} & \begin{tabular}[c]{@{}c@{}}MSCOCO,\\ Flickr-8k,\\ IAPR TC-12\end{tabular} &\begin{tabular}[c]{@{}c@{}}Reference\\ Model-based\end{tabular} & Gray & 2022 \\
\hline
\cite{ref41}$\ddagger$ & \begin{tabular}[c]{@{}c@{}}Cosine Similarity\\-Based\end{tabular} & \begin{tabular}[c]{@{}c@{}}OpenCLIP,\\ ViT-B/L,\\ ResNet\end{tabular} & \begin{tabular}[c]{@{}c@{}}LAION,\\ CC3/12M,\\ MSCOCO\end{tabular} & \begin{tabular}[c]{@{}c@{}}Target Model-\\based\end{tabular} & White & 2023 \\
\hline
\cite{ref42}$\ddagger$ & \begin{tabular}[c]{@{}c@{}}Representations-\\Based MIA\end{tabular} & \begin{tabular}[c]{@{}c@{}}Emotion\\Recognition\\Model\\(embedding\\sub-network)\end{tabular} & \begin{tabular}[c]{@{}c@{}}MuSE,\\ IEMOCAP,\\ MSP-Podcast,\\ MSP-Improv\end{tabular} & \begin{tabular}[c]{@{}c@{}}Target Model-\\based\end{tabular} & White & 2019 \\
\bottomrule
\end{tabular}
}
\begin{tablenotes}
\centering
\small
\item[1] \textit{$\ddagger$: Require access to target embeddings}
\end{tablenotes}
\end{table}

\begin{table}[hbt!]
\caption{Summary of Fine-Tuning Membership Inference Attacks against LMMs}
\label{table::finetune_mia_summary_2}
\centering
\resizebox{\textwidth}{!}{
\begin{tabular}{cccccccc}
\toprule
\textbf{Attack} & \textbf{Approach} & \textbf{Targets} & \textbf{FT-Method} & \textbf{Strategies} & \textbf{Scenario} & \textbf{Time} & \textbf{FT-Impact$^*$} \\
\hline
\cite{ref38} & LUMIA & \begin{tabular}[c]{@{}c@{}}LLaVa-\\OneVision\end{tabular} & -- & \begin{tabular}[c]{@{}c@{}}Target Model-\\based\end{tabular} & White & 2024 & N \\
\hline
\cite{ref55} & \begin{tabular}[c]{@{}c@{}}Rényi entropy-\\based\end{tabular} & \begin{tabular}[c]{@{}c@{}}miniGPT-4,\\ LLaVA,\\ LLaMA\\Adapter\end{tabular} & \begin{tabular}[c]{@{}c@{}}Partial FT,\\ Full FT,\\ Parameter-\\Efficient FT\end{tabular} & \begin{tabular}[c]{@{}c@{}}Target Model-\\based\end{tabular} & Black & 2024 & Y \\
\hline
\cite{ref43} & SMI & \begin{tabular}[c]{@{}c@{}}Pythia,\\ LLaVA\end{tabular} & -- & \begin{tabular}[c]{@{}c@{}}Target Model-\\based\end{tabular} & Black & 2025 & N \\
\hline
\cite{ref44} & \begin{tabular}[c]{@{}c@{}}Temperature-\\Based,\\Reference\\Inference,\\Image-only\\Inference\end{tabular} & \begin{tabular}[c]{@{}c@{}}LLaVA,\\ MiniGPT-4\end{tabular} & \begin{tabular}[c]{@{}c@{}}Instruction\\Tuning,\\ Partial\\FT\end{tabular} & \begin{tabular}[c]{@{}c@{}}Reference+Target\\ Model-based,\\Target Model-\\based,\\Target Model-\\based\end{tabular} & \begin{tabular}[c]{@{}c@{}}White,\\ Black,\\ Black\end{tabular} & 2025 & Y \\
\bottomrule
\end{tabular}
}
\begin{tablenotes}
\centering
\small
\item[1] \textit{--: Not Specified; $^*$: Analyzes the impact of the fine-tuning process and settings on vulnerability}
\end{tablenotes}
\end{table}

\subsubsection{\textbf{Visual Taxonomy by Adversarial Scenario, Strategy, and Target Model}}
{\bfseries \quad}

Unlike LLMs, which primarily encompass transformer-based NLP models, LMMs demonstrate significantly greater diversity due to their ability to process a combination of various input modalities. Such heterogeneity could also influence the corresponding MIA algorithm. Therefore, it is essential to also consider the type of target model, specifically the Vision Language Models (VLMs) and Multimodal Emotion Recognition (MER) models in this survey. 

A visual categorization of the analyzed MIA algorithms, organized by target model, adversary knowledge, and strategies, is presented in Figure 4 to foster reader comprehension and efficient navigation.

\begin{figure*}[tp]
  \centering
  \resizebox{0.95\linewidth}{!}{
    \begin{forest}
      for tree={
        grow=east,
        anchor=base west,
        parent anchor=east,
        child anchor=west,
        base=left,
        font=\small,
        rectangle,
        draw=black, % Changed from black!50 for a clearer border
        align=left,
        minimum width=2.5em,
        s sep=8pt,     % Increased vertical separation between siblings
        l sep=15pt,    % Added horizontal separation between levels
        inner xsep=3pt,
        inner ysep=2pt,
        edge path={
          % Corrected path: +10pt moves the line away from the parent node before turning
          \noexpand\path [\forestoption{edge}]
            (!u.parent anchor) -- +(10pt,0) |- (.child anchor) \forestoption{edge label};
        },
      },
      % Adjusted text widths to prevent overflow and improve layout
      where level=1{font=\scriptsize, text width=5em}{},
      where level=2{font=\scriptsize, text width=6em}{},
      where level=3{font=\tiny, text width=8.5em}{},
      where level=4{font=\tiny, text width=16em}{}, % Increased width for long content
      % Your original styles and tree structure follow
      ver/.style={rectangle, draw},
      leaf/.style={rectangle, draw},
      [MIA against LMMs
        [Strategy 
            [Reference Model[Temperature-Based MIA\cite{ref44}\\M${}^{4}$I MIA\cite{ref40}]]
            [Target Model[Rényi entropy-based MIA  \cite{ref55}\\Cosine Similarity-Based MIA\cite{ref41}\\LUMIA\cite{ref38}\\SMI\cite{ref43}\\Temperature-Based MIA\cite{ref44}\\Reference Inference MIA\cite{ref44}\\Image-only Inference MIA \cite{ref44}\\Representations-Based MIA\cite{ref42}]]
        ] 
        [Scenario
            [White-Box[LUMIA\cite{ref38}\\Temperature-Based MIA\cite{ref44}\\Cosine Similarity-Based MIA\cite{ref41}\\Representations-Based MIA\cite{ref42}]]
            [Gray-Box[M${}^{4}$I MIA\cite{ref40}\\SMI\cite{ref43}]]
            [Black-Box[Rényi entropy-based MIA\cite{ref55}\\Reference Inference MIA\cite{ref44}\\Image-only Inference MIA \cite{ref44}]]
        ]
        [Target
            [MER[Representations-Based MIA\cite{ref42}]]
            [VLM[M${}^{4}$I MIA\cite{ref40}\\Rényi entropy-based MIA  \cite{ref55}\\Cosine Similarity-Based MIA\cite{ref41}\\LUMIA\cite{ref38}\\SMI\cite{ref43}\\Temperature-Based MIA\cite{ref44}\\Reference Inference MIA\cite{ref44}\\Image-only Inference MIA \cite{ref44}]]
        ]
      ]
    \end{forest}
  }
  \caption{\textit{Taxonomy of MIA targeting LMMs}}
  \label{fig:llm_mia_taxonomy}
\end{figure*}

\section{Suggestion and Direction for Future Research}
{\bfseries Mitigating Dataset and Model Bias in MIAs} The study from M. Duan et al. \cite{ref54} demonstrates that the effectiveness of various MIAs, including MIN-K\% PROB \cite{ref28} and neighborhood comparison \cite{ref47} MIAs, is heavily influenced by the choice of dataset and target model. Furthermore, M. Meeus et al. \cite{ref68} highlight concerns with the post-hoc collection method, where non-member data are sampled from data sets published after model training. Such practices introduce temporal distribution shifts, meaning that apparent attack success may stem from stylistic differences rather than genuine memorization. These biases reduce the reliability of MIA results and obscure their implications for real-world deployments. Future research should systematically disentangle the influence of dataset bias and model bias on attack effectiveness, for example, by benchmarking across diverse datasets and model architectures, or by designing bias-aware calibration and normalization strategies. Such efforts would yield a more accurate understanding of MIA vulnerabilities and improve the generalizability of attack methodologies.

{\bfseries Threshold-Free Approaches for Black-Box MIAs} Many black-box MIAs, such as loss-based \cite{ref55}, perplexity-based \cite{ref29}, and MIN-K\% PROB \cite{ref28} MIAs, rely on manually defined thresholds to distinguish between member and non-member data. While thresholds provide a simple decision rule, their calibration is often dataset or model-specific, leading to performance variability and poor generalization across models and tasks. This dependence reduces the robustness of black-box MIAs and limits their applicability in real-world scenarios where calibration data may be unavailable. Future research should focus on developing and systematically evaluating threshold-free MIAs with techniques such as relative measures, statistical hypothesis testing, or adaptive calibration mechanisms, making MIAs more consistent, reliable, and transferable across large-scale models.

{\bfseries Addressing the Research Gap in MIAs against LMMs} Our survey reveals that research on MIAs against LMMs, particularly those beyond vision–language models, remains sparse and underdeveloped. Furthermore, existing studies focus primarily on pre-training or fine-tuning stages, while other critical parts of the pipeline, such as alignment or retrieval mechanisms, have received little to no attention. This gap leaves significant uncertainty about the privacy vulnerabilities of LMMs as they are increasingly deployed in diverse real-world applications. Future research should therefore expand beyond early-stage evaluations to develop advanced MIA algorithms tailored to the unique structures and development pipelines of LMMs. Establishing systematic methodologies in this direction would provide a more comprehensive understanding of privacy risks in multimodal systems and support the design of more resilient defenses.

{\bfseries Prioritizing Privacy-Critical Data in MIAs} Most existing MIA studies evaluate privacy leakage by treating all training samples equally and measuring aggregate exposure. However, not all training data points are equally vulnerable to MIA. Current protection mechanisms, such as Differential Privacy (DP), provide strong privacy guarantees but often incur high computational costs and noticeable performance degradation when applied uniformly across all data. This mismatch between generalized protection and differentiated sensitivity limits their practical adoption. Future research should aim to identify which subsets of training data are more privacy-critical and design selective protection mechanisms that prioritize safeguarding these high-risk samples. Such an approach would enhance privacy resilience while minimizing unnecessary performance trade-offs, thereby bridging the gap between privacy guarantees and real-world needs.

{\bfseries Improving the Transferability of MIAs Across Models} Some existing MIA algorithms are designed with strong model dependencies, limiting their applicability to specific categories. For example, cosine similarity–based MIAs \cite{ref41} are tailored to VLMs, while attention-based MIAs \cite{ref48} are mainly applicable to LLMs or LMMs that adopt LLM backbones. Such specialization reduces the practical utility of these approaches, as their effectiveness does not easily transfer across different architectures or modalities. Future research should focus on developing cross-model MIA techniques that exploit shared vulnerabilities across architectures, like perplexity-based MIA \cite{ref29}, or incorporate adaptable components, such as SMI \cite{ref43}, enabling attacks to remain effective in heterogeneous environments.

{\bfseries Architectural Factors Shaping MIA Vulnerability in LMMs} LMMs are inherently flexible, as each of their core components can be implemented with diverse algorithmic choices. This modularity introduces unique and heterogeneous sources of privacy risk. Key factors that may influence MIA vulnerability include the type of unimodal sub-models adopted for encoding different modalities, the architecture of the cross-modal learning network, the distribution of trainable parameters across components, and other design characteristics. Understanding how these architectural and training choices affect membership leakage is essential for moving beyond post-hoc protection and towards privacy-aware design. Preliminary evidence suggests that certain design choices amplify privacy risks. For example, the study of temperature-based \cite{ref44} and Rényi entropy-based \cite{ref55} MIAs both indicate that the vulnerability of fine-tuned LMMs increases with the number of trainable parameters involved in fine-tuning. However, the literature remains sparse, and a systematic analysis is still lacking. Future work should undertake a comprehensive analysis of these factors to optimize privacy resilience, enabling the development of LMMs that are robust by design while reducing reliance on costly or performance-degrading post-training defense mechanisms.

{\bfseries Disentangling Fine-Tuning Membership in MIAs} While several recent works have begun to explore MIAs in fine-tuned large-scale models, a key challenge remains overlooked: both pre-training and fine-tuning datasets are technically members of the final model. Thus, identifying whether a data point belongs specifically to the fine-tuning dataset, which is typically more privacy-sensitive, requires not only confirming its membership but also distinguishing it from the pre-training dataset. An intuitive but limited approach would be to reapply MIAs to the original pre-trained model for comparison, but this is infeasible under black-box settings. Moreover, various studies have shown that the fine-tuning process alters the overall MIA vulnerability of the model \cite{ref32,ref49,ref50}, making the pre-trained model less suitable as a calibration baseline. Future research should therefore develop novel attack strategies capable of disentangling fine-tuning membership from pre-training membership even without relying on the original model. Advancing this direction would strengthen the robustness of fine-tuning MIAs and provide a clearer understanding of privacy risks in real-world deployment pipelines.

{\bfseries Towards a Unified Benchmark for MIAs and Defenses} Current MIA studies are often evaluated under fragmented conditions, with each work relying on different datasets, models, assumptions, and baselines. This lack of standardization makes it difficult to compare results across studies and obscures conclusions about the generalizability and robustness of both attacks and defenses. As a result, progress in the field is less efficient, since it remains unclear whether improvements are due to methodological advances or differences in experimental settings. Future research should prioritize the development of a unified benchmarking framework for MIAs and their defenses, incorporating diverse model families such as LLMs, LMMs, and other machine learning architectures. Such a framework would enable systematic, reproducible, and fair evaluations across hybrid targets, facilitating more rigorous comparisons and accelerating the identification of both effective attacks and resilient defense mechanisms.

\section{Conclusion}

With the rapid advancement of technology, large-scale machine learning models, including LLMs and LMMs, are increasingly being deployed across various domains. In this work, we covered most existing studies about MIA against LLMs and LMMs, systematically reviewing each study, categorizing different MIA approaches, and analyzing their strengths and limitations. Furthermore, given the widespread adoption of fine-tuning and other pipelines in model development and deployment, we also examined current MIA against LLMs and LMMs across pipelines. Based on the research gaps identified in the survey, we proposed several directions for future research, aiming to improve the practicality, effectiveness, and robustness of MIAs in real-world scenarios. We hope that this work serves as a valuable resource for researchers and developers working in machine learning security and privacy, fostering further advancements in the field.

\bibliographystyle{ACM-Reference-Format}
\bibliography{reference.bib}
%%\bibliography{sample-base}

\end{document}